\documentclass[pdflatex,sn-mathphys]{sn-jnl}

\usepackage{subfigure}
\usepackage{natbib}
\usepackage{cleveref}
\usepackage{lineno}

\jyear{2021}%

\theoremstyle{thmstyleone}%
\theoremstyle{thmstyletwo}%

\theoremstyle{thmstylethree}%

\begin{document}

\title[ ]{DLP-GAN: learning to draw modern Chinese landscape photos with generative adversarial network}


\author[1]{\fnm{Xiangquan} \sur{Gui}}

\author*[1]{\fnm{Binxuan} \sur{Zhang}}\email{212085400201@lut.edu.cn}

\author[2]{\fnm{Li} \sur{Li}}
\author[2]{\fnm{Yi} \sur{Yang}}

\affil[1]{\orgdiv{School of Computer and Communication}, \orgname{Lanzhou University of Technology}, \orgaddress{ \city{Lanzhou}, \postcode{730050}, \state{Gansu}, \country{PR China}}}

            
\affil[2]{\orgdiv{School of Information Science and Engineering}, \orgname{Lanzhou University}, \orgaddress{ \city{Lanzhou}, \postcode{730000}, \state{Gansu}, \country{PR China}}}


\abstract{Chinese landscape painting has a unique and artistic style, and its drawing technique is highly abstract in both the use of color and the realistic representation of objects. Previous methods focus on transferring from modern photos to ancient ink paintings. However, little attention has been paid to translating landscape paintings into modern photos. To solve such problems, in this paper, we (1) propose DLP-GAN (\textbf{D}raw Modern Chinese \textbf{L}andscape \textbf{P}hotos with \textbf{G}enerative \textbf{A}dversarial \textbf{N}etwork), an unsupervised cross-domain image translation framework with a novel asymmetric cycle mapping, and (2) introduce a generator based on a dense-fusion module to match different translation directions. Moreover, a dual-consistency loss is proposed to balance the realism and abstraction of model painting. In this way, our model can draw landscape photos and sketches in the modern sense. Finally, based on our collection of modern landscape and sketch datasets, we compare the images generated by our model with other benchmarks. Extensive experiments including user studies show that our model outperforms state-of-the-art methods.}




\maketitle

\section{Introduction}\label{sec1}

Landscape painting is a significant subject in Chinese art. Brush and ink, unique to this style of painting, are considered the soul of Chinese landscape painting. By adjusting moisture, ink color, and brush strokes, artists can create endless variations in their landscapes. However, mastering these techniques takes time and requires a high level of skill, which can be a barrier to the development of this art form \citep{liu2021basic}. Therefore, we are exploring the use of modern technology to create landscape photos that capture the moods and emotions of ancient Chinese landscapes.

\begin{figure}[ht]
\centering
\includegraphics[width=\textwidth]{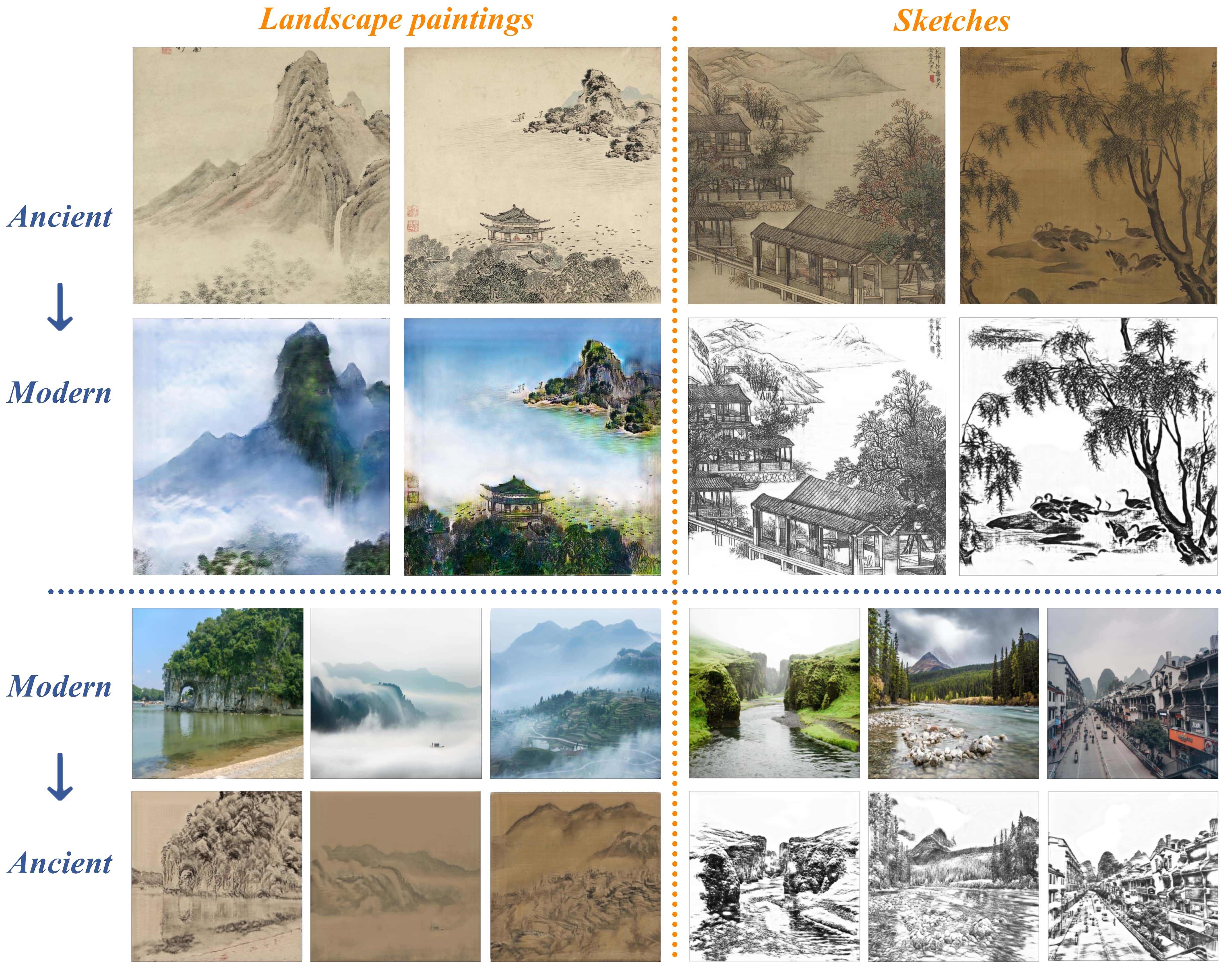}
\caption{Example results of DLP-GAN in landscape painting style transfer. Top: ancient landscape painting to modern photos. Bottom: modern photos to ancient landscape painting. Left: landscape painting. Right: sketches. }
\label{figshow}
\end{figure}

Computer-assisted art creation has greatly expanded the ability of artists to create their own works. With the rise of artificial intelligence, natural images can be transformed using techniques such as style transfer \citep{li2017,gatys2016,johnson2016} and image-to-image translation \citep{zhu2017unpaired,zhu2017toward,isola2017}. However, most research in this area has focused on oil and watercolor paintings \citep{li2020sdp,lin2021drafting,liu2021adaattn}, leaving a gap in the ability to apply these techniques to landscape paintings. Our work addresses this gap by exploring the novel asymmetric framework DLP-GAN to transfer the style of Chinese landscape paintings.

The creation of paintings exhibits significant cultural differences between the East and the West. Chinese paintings are not bound to the reality of the object, but rather to the subjective truth behind it \citep{liu2021basic}. Due to such differences, researchers have studied Chinese painting style transfer using deep convolutional networks to extract contours and learning various drawing techniques through adversarial learning\citep{peng2022contour,zheng2018,zhou2019}. However, these methods are less efficient because they require two stages to complete. Another suitable method is unsupervised cross-domain image-to-image translation, where the key is to utilize generative adversarial networks \citep{goodfellow2014}. Unlike the generation tasks of the previous two stages, the GAN-based method has the benefit of capturing high-level styles from a set of paintings. This process is mainly implemented by the generator, which learns the mapping between two domains until it generates images that the discriminator cannot distinguish from real paintings. To capture image features, common solutions include (i) training cross-domain translation mappings based on cycle-consistency constraints \citep{bharti2022emocgan,he2018,wang2022}, (ii) learning end-to-end style transfer by matching different cross-domain loss functions \citep{li2018neural,qiao2019,qin2022towards}, and (iii) separating the characteristic representations of style and content to achieve style transfer in landscape painting by re-encoding \citep{sun2022,li2021,xue2021}.
In addition to GANs, diffusion models have demonstrated their potential in generating high-quality images \citep{dhariwal2021diffusion,ho2020denoising}, and several conditional diffusion models \citep{saharia2022palette,su2022dual,li2023bbdm} have been proposed for image-to-image translation tasks. However, these models often exhibit limited generalization capabilities, particularly in specific applications.

\begin{figure}[ht]
\centering
\includegraphics[width=\textwidth]{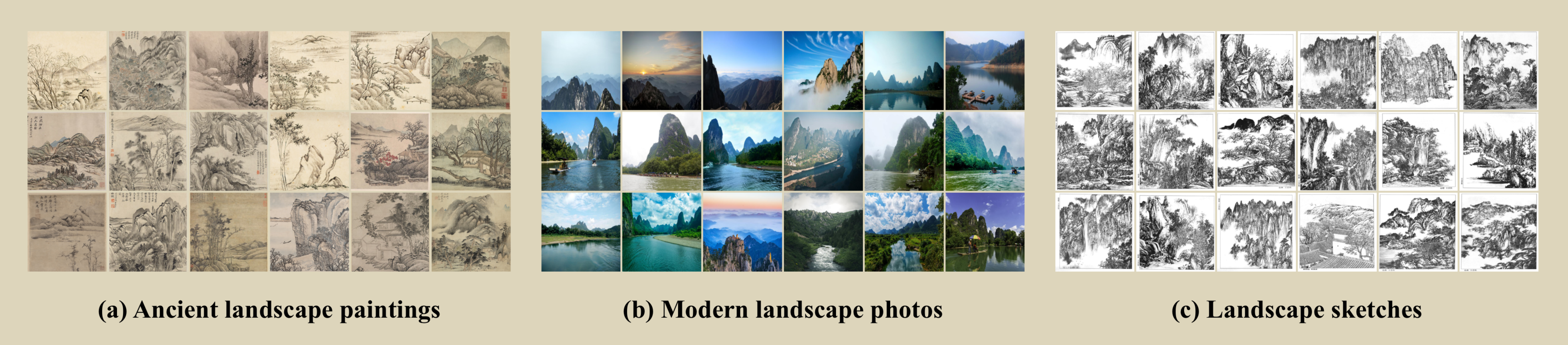}
\caption{Some example pictures collected for our style transfer tasks. (a) Ancient landscape paintings for ancient $\rightarrow$ modern. (b) Modern landscape photos for modern $\rightarrow$ ancient. (c) Landscape sketches for ancient $\rightarrow$ line drawing.}
\label{fig2}
\end{figure}

Although the visual quality of the results generated by such methods has improved, the style effect is still not obvious in images containing backgrounds with complex elements, such as the sky and trees in the background of ancient landscape paintings. The transfer from ancient paintings to modern photos has not been studied. Based on this, our study focuses on transferring ancient landscape paintings (see \cref{fig2}-a) to modern photos (see \cref{fig2}-b) and sketches (see \cref{fig2}-c). Due to the huge domain deviation, generating different image types also poses different challenges. For modern landscape photos, the distribution of target styles is diverse because they have relatively large variations in texture, color, and brush stroke size, which makes it difficult to capture the exact modern style. Thus, we need a deep convolutional network with robust generative capability during the translation. For landscape sketches, which have a single target style distribution consisting of edges and shadows drawn in dark lines, capturing their styles is not as difficult as in modern photos. The main challenge in producing satisfactory results is to accurately identify and preserve edge features in landscape sketches.

We propose DLP-GAN to overcome these challenges. In our framework, a dense-fusion module-based generator is designed to translate landscape paintings into realistic photos. The dense-fusion module \citep{li2018dense} has the advantage of enhancing feature propagation and encouraging feature reuse over the residual blocks used in CycleGAN \citep{zhu2017unpaired}. To further simulate the features highlighting the edge strokes of the drawing subject, we propose a semantic-consistency loss to learn the salient edge stroke styles. This loss function enhances the accuracy of cross-domain image translation by constraining the shared semantic encoding between the input image and the reconstructed image. We also improve the model using a feature-consistency loss, which maximizes the abstract representation of drawing images by constraining the feature space of the source domain to overlap with the target domain. Our approach introduces a dual-consistency loss that combines both feature and semantic consistency components, effectively addressing the challenges of capturing style and preserving content in image translation tasks. By adjusting the weights of these components, a balance between realism and abstraction can be achieved. Several examples of landscape paintings generated using our approach are shown in \cref{figshow}. Furthermore, we collect a high-resolution dataset of modern landscape photos and their sketches to advance the field.

Our model eventually learns to draw modern landscape photos and sketches. In short, the main contributions of our work are as follows: (i) We introduce DLP-GAN, a novel approach aiming to translate ancient landscape paintings into modern image domains. We design a generator based on a dense-fusion module to learn stroke features, which performs better in cross-domain image translation. (ii) We propose an unsupervised image translation framework with unpaired datasets, which is valuable for fields in which paired datasets are difficult to collect. (iii) We introduce a novel dual-consistency loss, which combines feature and semantic consistency components. This loss function effectively addresses the challenges of capturing style while preserving the essential content in image translation tasks. Extensive experiments show that our method generates Modern landscape photos with realistic effects.

This paper is structured as follows: In \cref{sec2}, we review the related work. We describe the proposed method in \cref{sec3}. In \cref{sec4}, We report the results of qualitative and quantitative experiments. Then, in \cref{sec5}, a conclusion and future work are presented.

\section{Related work}\label{sec2}

\subsection{GAN for image-to-image translation}

Goodfellow et al.\citep{goodfellow2014} proposed a new framework for generative adversarial network (GAN), which can be used to estimate generative models through an adversarial process. The generator and discriminator are trained in parallel, with the generator learning the distribution of real data and the discriminator learning to distinguish between real and fake data. GAN have been rapidly extended to many tasks, and have achieved notable success in the area of image-to-image translation.

Image-to-image translation tasks can be divided into supervised and unsupervised types. Supervised translation uses class labels or source images as conditions to generate high-quality images, which can further be divided into directed and bi-directional translation. Pix2Pix \citep{isola2017}, a representative of directed translation, is a supervised method that learns one-to-one mappings. However, the images generated by this method are still limited by low resolution and blurriness. To address this issue, Wang et al.\citep{wang2018} proposed Pix2pixHD, which increases the resolution of the output image to 2048 $\times$ 1024. BicycleGAN \citep{zhu2017toward}, a representative of bi-directional translation, is a multi-model cross-domain translation method that requires pairs of training images. It combines a Conditional Variational Autoencoder (CVAE) and Conditional Latent Regression (CLR) to generate realistic and diverse outputs. Unsupervised approaches, on the other hand, aim to learn the mapping between two domains without the need for paired data. CycleGAN \citep{zhu2017unpaired} proposes a novel cycle consistency constraint to build a bi-directional relationship between two domains. To capture the features of variables in the domain, representation deconstruction has been investigated in MUNIT \citep{huang2018multimodal}, where content and style encoders are used to extract features of content and style, respectively. However, these methods often have difficulty building robust and balanced representations for specific domains, making them less adaptable to various translation tasks and prone to failure in some cases. To overcome this limitation, we propose a dense-fusion module-based generator, a dual-consistency loss to learn stroke features, and apply it to image translation tasks to generate high-quality images.

\subsection{Style transfer for ancient Chinese painting}

With the increasing popularity of deep neural networks, many researchers have been focusing on style transfer for traditional Chinese paintings. Zheng and Zhang\citep{zheng2018} propose using multi-scale neural networks for Chinese image transfer, but their method is not end-to-end and requires sketches or edges as inputs. Li et al.\citep{li2018neural} propose a neural network-based method that learns end-to-end transfer through a novel filter and differentiable loss terms. Qiao et al.\citep{qiao2019} propose a Domain Style Transfer Network (DSTN) for transferring ancient paintings to realistic images, but this method focuses on ancient Chinese flower and bird landscape paintings. Recent methods based on end-to-end generative adversarial network architectures explore the translation from photos to Chinese ink painting style, such as \citep{zhang2020,he2018}. Li et al.\citep{li2021} train a fast feed-forward generative network to extract the corresponding styles that can be transferred from photographs to traditional Chinese portraits. Xue\citep{xue2021} proposes a two-stage architecture for generating Chinese landscape paintings without conditional input. Recently, Chung and Huang\citep{chung2022} generate more accurate real images by enhancing the details of boundary images, but their results require user interaction. Wang et al.\citep{wang2022} use an improved asymmetric CycleGAN model to solve the transfer from real photos to ink painting style. Unlike these methods, our proposed method focuses on generating modern landscape photos and sketches using ancient landscape paintings as inputs. Our proposed DLP-GAN model uses a higher-resolution image and stricter constraints for optimization to handle various tasks more effectively.

\section{Method}\label{sec3}

\subsection{Overview }
In this section, we propose a GAN-based model with a novel asymmetric cycle structure that transforms ancient landscape paintings into high-quality modern photos. Our model uses unpaired training data and has the advantages of being end-to-end and requiring no user interaction. Let X and Y be the ancient landscape painting domain and the modern photo domain. Our model learns a function $\Phi$ that maps from X to Y using training data $\mathcal{N}(x) = \{x_{i} \mid i=1,2, \cdots, n_{i}\} \subset \text{X}$ and $\mathcal{N}(y) = \{y_{j} \mid j=1,2, \cdots, n_{j}\} \subset \text{Y}$. $n_{i}$ and $n_{j}$ are the numbers of training set images in the X and Y domains, respectively. 

\begin{figure}[ht]
\centering
\subfigure{
\begin{minipage}[b]{\textwidth}
\includegraphics[width=\textwidth]{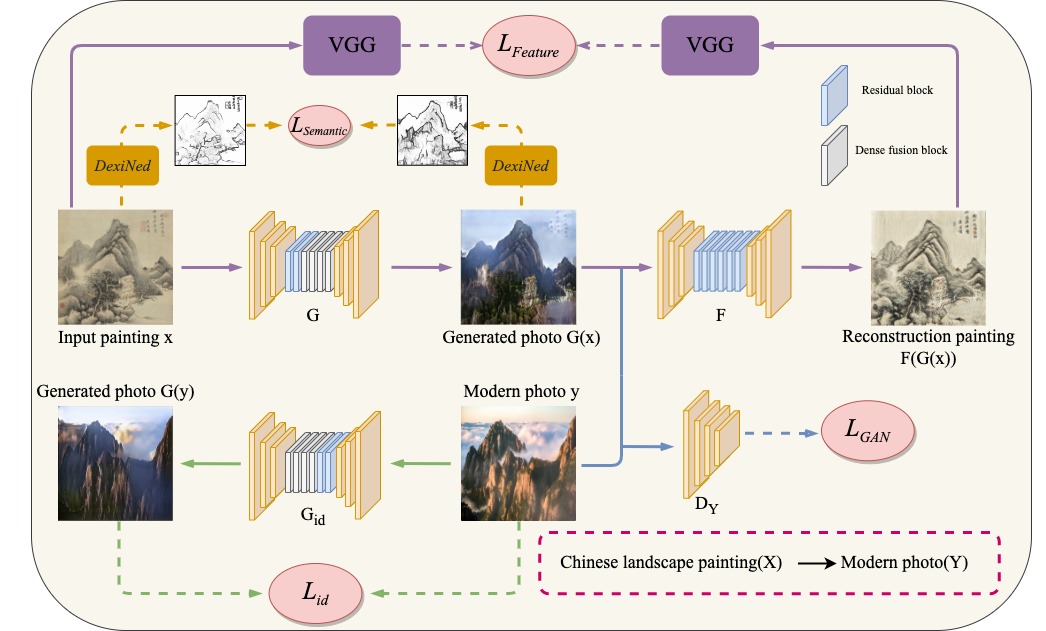}
\end{minipage}}\\
\subfigure{
\begin{minipage}[b]{\textwidth}
\includegraphics[width=\textwidth]{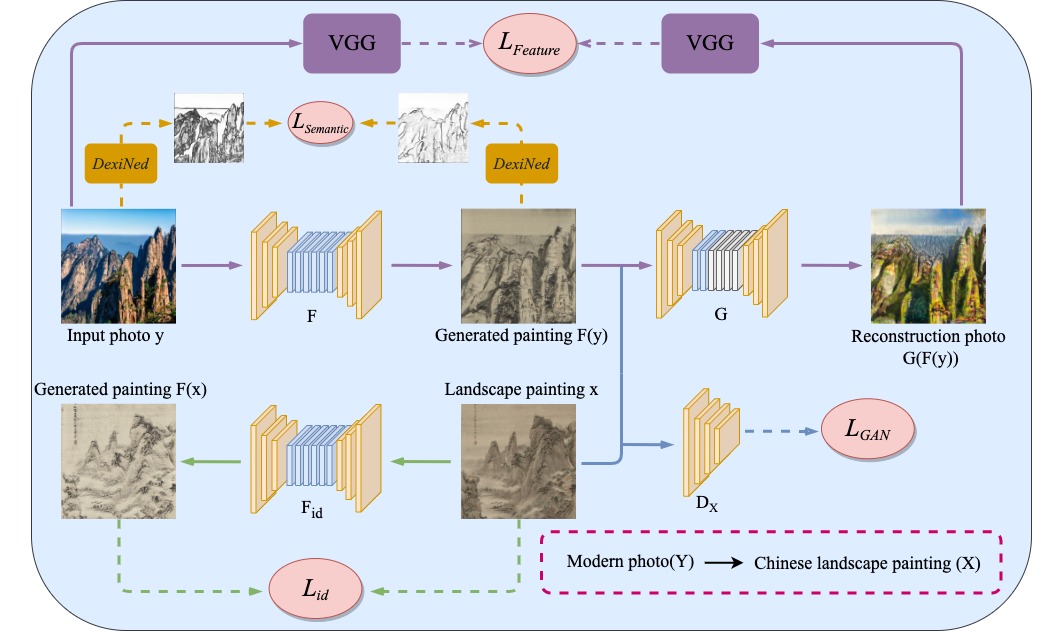}
\end{minipage}}
\caption{The pipeline exhibits the architecture of the proposed DLP-GAN. The pipeline consists of two directions operating different generators. Top: translation direction from Domain (X) Chinese landscape painting $\rightarrow$ Domain (Y) Modern photos. Bottom: Opposite direction.}
\label{fig3}
\end{figure}

The overall framework of the proposed model is shown in \cref{fig3}, which contains pipelines in two directions, from the ancient landscape painting X domain to the modern photo Y domain direction (X $\rightarrow$ Y) and the opposite direction (Y $\rightarrow$ X). Each pipeline includes the following modules: generator $G$, generator $F$, discriminator $D$, $DexiNed$ module for edge detection, and a pre-trained $VGG$ network module for feature extraction. Generators $G$ and $F$ and discriminator $D$ work together to learn the mapping between domain X and Y.

There is much more information in the modern photo domain than in the ancient landscape painting domain. For example, modern photos often contain a variety of colors and details in the background, such as the colors of the sky, trees, and buildings. In contrast, the background of ancient landscape paintings is often left blank or minimally detailed. As shown in the CycleGAN method in \cref{fig7}, if the same cycle-consistency loss is applied to the reconstructed and input photos, it can cause the generated images to be distorted and limit the ability of the model to create meaningful mappings between the two domains.

To solve the problem of distorted generated images and limited mapping capabilities, our proposed model uses an asymmetric structure with a strict generator $G$ in the (X $\rightarrow$ Y) direction and a relaxed generator $F$ in the opposite direction. The strict generator $G$ focuses on accurately capturing the specific characteristics and details of the target domain, while the relaxed generator $F$ aims to preserve the overall style and essence of the source domain. This allows for more flexible optimization of the objective function and enables the model to focus on the important features in modern photos. In addition, we propose a dual-consistency loss function that forces the embedded information to be visible in the generated images, and an identity loss function is added to regularize the generators and avoid meaningless translations. By effectively combining these techniques, our model generates high-quality modern photos and sketches, preserving spatial features and enhancing the fidelity of elemental details in the input images.

The network in our model is trained using an adversarial process, where the two discriminators ($D_{X}$ and $D_{Y}$) engage in a competitive interplay with the two generators (\emph{G} and \emph{F}) until they reach a point of equilibrium. This equilibrium is commonly referred to as the Nash equilibrium. The total loss function $L_{\text {Total}}\left(G, F, D_{X}, D_{Y} \right)$ for the model includes five types of loss terms: generate adversarial loss 
$L_{LSGAN}\left(G, D_Y \right)$ and $L_{LSGAN}\left(F, D_X \right)$, dual-consistency loss $L_{\text {Dual}}$ including feature loss $L_{\text {Feature}}(G, F)$ and semantic loss $L_{\text {Semantic}}(G, F)$, and identity loss $L_{i d}(G, F)$. The function $\Phi$ is then optimized through the solution of a specific minimax problem using a defined loss function:

\begin{equation}\label{equ1_1}
\begin{aligned}
    \min _{G, F} \max _{D_{X}, D_{Y}} & L_{\text {Total}}\left(G, F, D_{X}, D_{Y}\right)\\
    &=\lambda_{G A N} L_{G A N}+\lambda_{D u a l} L_{D u a l}+\lambda_{i d} L_{i d}\\
    &=\lambda_{G A N} \Big(L_{LSG A N}\left(G, D_Y \right)+L_{LSG A N}\left(F, D_X \right)\Big)\\
    &+\lambda_{D u a l}\Big(L_{\text {Feature}}(G, F)+\mu L_{\text {Semantic}}(G, F)\Big)\\
    &+\lambda_{i d} L_{i d}(G, F)
\end{aligned}
\end{equation}

Where the weights $\lambda_{\text {GAN }}$, $\lambda_{\text {Dual }}$, $\lambda_{\text {id }}$ control the relative importance of the component losses to the overall learning objective.

The network architectures for \emph{G}, \emph{F}, $D_{X}$, and $D_{Y}$ are introduced in \cref{sub_net}. The detailed design of several relevant objective functions are presented in \cref{sub_fun}. The architecture of the proposed DLP-GAN is illustrated in \cref{fig3}.

\subsection{Network architectures }\label{sub_net}

In the (X$\rightarrow$Y) direction, as shown at the top of \cref{fig3}, the framework is given two semantically related stylized image domains (X, Y). First, the ancient landscape image x is input to generator $G$ and output as translated image $G(x)$, then generator $F$ encodes the image $G(x)$ into a reconstructed image $F(G(x))$. To maintain color space consistency within the same domain, we have adopted an approach inspired by the identity loss in CycleGAN \citep{zhu2017unpaired}. This adaptation ensures consistent color spaces between input and output within a given domain. In the opposite direction, as shown at the bottom of \cref{fig3}. 

The DLP-GAN is an asymmetry-based framework, where the generator is an encoder-decoder, and there is a transformer network based on the residual structure in the middle; the transformer network determines the representation capability of the generator, and we use two versions of the generator network $G$ and $F$.

\begin{figure}[ht]
\centering
\includegraphics[width=0.8\textwidth]{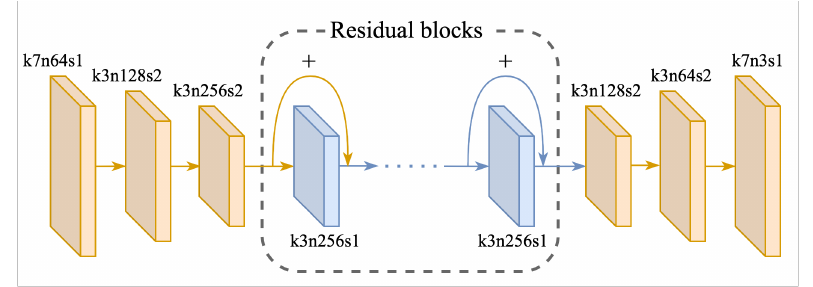}
\caption{Architecture of the generator $F$, in which k represents the kernel size, n is the number of feature maps and s is the stride in each convolutional layer.}
\label{fig4}
\end{figure}
\begin{figure}[ht]
\centering
\includegraphics[width=0.95\textwidth]{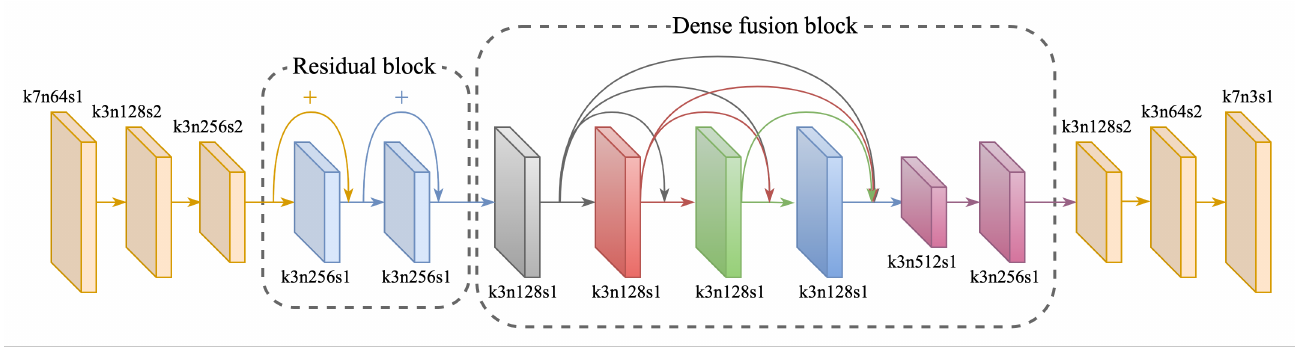}
\caption{Architecture of the generator $G$.}
\label{fig5}
\end{figure}

\textbf{The generator $F$} structure is shown in \cref{fig4}. The encoder consists of a 7 $\times$ 7 convolutional layer, followed by two downsampling with stride 2. The intermediate structure includes six residual blocks, each of which contains two convolutional layers, and the decoder consists of two transposed convolutional blocks with stride 2 for upsampling, and one 7 $\times$ 7 convolutional layer.

\textbf{The generator $G$} structure is shown in \cref{fig5}. The intermediate structure starts with two residual blocks. We additionally embed a dense-fusion block, including a dense block and a fusion block. The fusion block fuses the information from each previous layer, and by this operation, our network can retain more useful information from the intermediate layers and is easy to train. This design not only enhances the accuracy of the drawing results but also improves overall efficiency.

In traditional CNN-based networks, the issue of degradation arises with increased network depth, limiting the utilization of information extracted by intermediate layers \citep{he2016deep}. To overcome this, we draw inspiration from the concept of dense blocks introduced by \citep{huang2017densely}, which establish direct connections from any layer to all subsequent layers. This architecture offers several advantages, including preserving maximum information throughout the network, improving information flow and gradient propagation for easier training, and providing a regularizing effect that reduces overfitting.

\begin{figure}[ht]
\centering
\includegraphics[width=\textwidth]{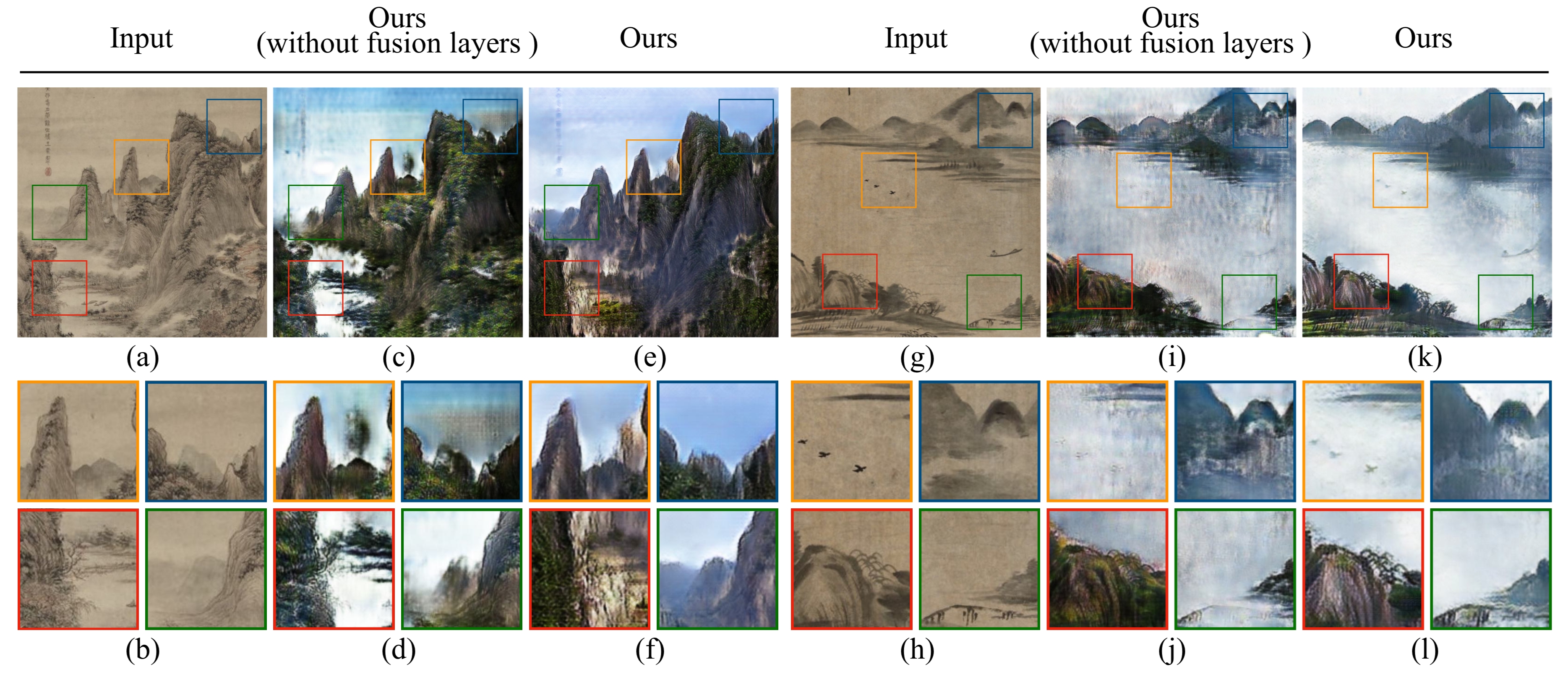}
\caption{Comparison of image details generated by our method with and without fusion layers. While the images produced by the two methods look similar overall, the images produced by the model after adding the fusion layer look aesthetically better and more detailed. }
\label{fignew}
\end{figure}

Based on these observations, our proposed model embeds a dense-fusion module behind the residual blocks in the generator network. \Cref{fignew} presents a comparison of the generated image details with and without fusion layers. Although the overall appearance of the images produced by both methods may appear similar, the inclusion of fusion layers enhances the aesthetic appeal and level of detail in the generated images. The images generated by the model with fusion layers exhibit a superior visual quality, capturing finer elements and nuances with greater fidelity. This architectural choice enables the model to preserve artistic style while effectively addressing discontinuities at semantic edges.

$\textbf {The discriminator } D_{X}$, and $\textbf {discriminator } D_{Y}$ represent discriminators in the style domains X and Y, respectively. The main function of the discriminator is to distinguish the generated photos from the real ones. In this paper, we utilize the Patch GAN discriminator introduced by Isola et al. \citep{isola2017}. The Patch GAN discriminator offers several advantages, including reduced number of parameters, faster computation, and the ability to handle images of arbitrary sizes. The discriminator $D_{X}$, and $ D_{Y}$ networks consist of four down-sampling convolutional layers, each followed by an additional instance normalization layer and a $ReLU$ nonlinear activation function. The last layer is a convolutional layer with a kernel size of 4 and a stride of 1.

\subsection{Loss functions}\label{sub_fun}

There are five types of losses in our loss function (Eq.\ref{equ1_1}). We explain them in detail as follows.

\textbf{Generate adversarial loss} $L_{G A N}$ is used to optimize the generator and the discriminator networks in the proposed model. The generator-discriminator pairs are denoted as ($G, D_{Y}$) and ($F, D_{X}$), and each generator is trained jointly with its corresponding discriminator. The $L_{G A N}$ loss function helps the generated image to be closer to modern photos during the translation process. During training, the adversarial network learns two mapping functions between two directions: $(G, D_Y)$ and $(F, D_X)$. The $L_{G A N}$ loss function in the (X $\rightarrow$ Y) direction, which is composed of the generator $G$ and the corresponding discriminator $D_{Y}$, is defined as follows:

\begin{equation}
\begin{aligned}
    L_{G A N}\left(G, D_Y\right)&=\mathrm{E}_{y \sim p_{\text {data }}(y)}\Big[\log D_Y(y)\Big]\\
    &+\mathrm{E}_{x \sim p_{\text {data }}(x)}\Big[\log \big(1-D_Y(G(x))\big)\Big] \label{equ1}
\end{aligned}
\end{equation}

Where $p_{\text {data }}(x)$ and $ p_{\text {data }}(y) $ represent the data distribution in the X and Y domains, respectively. Similarly, we introduce similar adversarial loss $L_{G A N}\left(F, D_X\right) $ in the (Y $\rightarrow$ X) direction, which consists of the generator $G$ and the corresponding discriminator $D_{X}$.

In our training process, we substituted the conventional negative log-likelihood loss used in regular GANs \cite{goodfellow2014} with the least squares loss introduced by Mao et al. \cite{mao2017}, referred to as Least Squares GANs (LSGANs). For the $L_{GAN}\left(G, D_Y\right)$, the adversarial loss is defined as:

\begin{equation}
L_{L S G A N}\left(D_Y\right) =\frac{1}{2} E_{y \sim p_{\text {data }}(y)}\Big[D_Y(y)-1\Big]^{2} 
+\frac{1}{2} E_{x \sim p_{\text {data }}(x)}\Big[D_Y\big(G(x)\big)\Big]^{2}
\label{equ2}
\end{equation}
\begin{equation}
    L_{L S G A N}(G) =\frac{1}{2} E_{x \sim p_{\text {data }}(x)}\Big[D_Y\big(G(x)\big)-1\Big]^{2}
\end{equation}

Similarly, for the GAN loss $L_{G A N}\left(F, D_X\right) $ the adversarial loss is defined as:

\begin{equation}
L_{L S G A N}\left(D_X\right) =\frac{1}{2} E_{x \sim p_{\text {data }}(x)}\Big[D_X(x)-1\Big]^2 
+\frac{1}{2} E_{y \sim p_{\text {data }}(y)}\Big[D_X\big(F(y)\big)\Big]^2
\end{equation}
\begin{equation}
L_{L S G A N}(F)=\frac{1}{2} E_{y \sim p_{\mathrm{data}}(y)}\Big[D_{X}\big(F(y)\big)-1\Big]^{2}
\end{equation}

\textbf{Dual-consistency loss} $L_{ \text { Dual }}$ is used to perform high-quality image translation between two style domains. The main idea is to constrain the semantic consistency of the input image with the generated image and the feature consistency with the reconstructed image. The goal is to learn the mapping from the input to the generated image domain with high quality. It can be specifically divided into feature loss and semantic loss, as defined below.

\textbf{Feature loss} $L_{ \text { Feature}}$ aims to make the output more creative by learning additional styles while preserving the semantics of the generated image. The most straightforward way to preserve the desired information from the content images is to use image reconstruction loss \citep{isola2017,zhu2017unpaired}, which is calculated at the pixel-level.
However, since our task is to switch from information-poor to information-rich domains, pixel-level loss is not conducive to the creativity of the model. It has been found that pre-trained VGG networks on Image Net \citep{simonyan2014} can extract high-level semantic information, so we use images extracted from the pre-trained VGG-16 as feature representations and evaluate the similarity between the input and output images at a higher level of abstraction. The feature-consistency loss $L_{\text {Feature }}$ is formulated as:

\begin{equation}
\begin{aligned}
L_{\text {Feature }}(G, F) &= E_{x \sim p_{\text {data }}(x)}\Big[\big\|\text{VGG}_{\text {relu3\_3 }}(x)- \text{VGG}_{\text{relu3\_3}}F\big(G(x)\big)\big\|_1\Big]\\
&+E_{y \sim p_{\text {data }}(y)}\Big[\big\|\text{VGG}_{\text {relu3\_3 }}(y)- \text{VGG}_{\text{relu3\_3}}G\big(F(y)\big)\big\|_1\Big]
\end{aligned}
\end{equation}
$V G G_{\text {relu3\_3 }}$ is the feature map extracted from the ${\text {relu3\_3 }}$ layer of the VGG-16 networks.

To achieve conversion between two style domains and perform better semantic style transfer, we hope the generator network can learn the semantic mapping relationship between the two domains related to the topic. Therefore, we propose semantic-consistency loss to train the generator network.

\textbf{Semantic loss} $L_{ \text { Semantic }}$ mainly simulates the semantic style that emphasizes the subject in the scene, which is obtained using the edge of the salient subject. As mentioned earlier, there is much less information in the domain Y than in the domain X. We do not expect the results of $x$ $\rightarrow$ $G(x)$ and $y$ $\rightarrow$ $F(y)$ to be similar in pixels. Instead, we only expect the information in the edge semantics of $x$ and $G(x)$ to be similar. The DexiNed network can extract multiple semantic levels for different types of brushes and strokes. We use a pre-trained DexiNed network \citep{poma2020} to extract edge semantic content from the input and translated images, and then evaluate the similarity of edge semantics using the $L_{lpips}$ perceptual metric proposed by Zhang et al.\citep{zhang2018}. In this way, the generator can better encode landscape paintings and extract useful features. The semantic-consistency loss $L_{\text {Semantic }}$ is defined as:

\begin{equation}
\begin{split}
L_{\text {Semantic }}(G, F) &= E_{x \sim P_{\text {data }}(x)}\Big[L_{\text {lpips }}\big(Dex\left(x\right),Dex\big(G\left(x\right)\big)\big)\Big]\\
&+E_{y \sim P_{\text {data }}(y)}\Big[L_{\text {lpips }}\big(Dex\left(y\right),Dex\big(F\left(y\right)\big)\big)\Big]\\
\end{split}
\end{equation}
Dex represents the pre-trained DexiNed network. 

The above two losses constitute a dual-consistency loss. By minimizing the semantic loss and the feature loss, the generator network can preserve the semantic content and the style information in the original image. In other words, the dual-consistency loss constrains the generator network from learning semantic style-related image mapping relations between two style domains, and the dual-consistency loss $L_{\text {Dual }}$ is then defined as:

\begin{equation}\label{equ9}
    L_{ \text { Dual }}(G, F) = L_{\text {Feature }}(G, F)+\mu L_{\text {Semantic }}(G, F)
\end{equation}
Where $\mu$ is the weight parameter used to balance the feature and semantic correlation of the generated image, $\mu$ = 1 in this paper, and the meaning of the $\mu$ parameter will be described in detail later.

\textbf{Identity loss} $L_{i d}$ is a regularization term that is used to avoid producing nonsensical translations with a generator. As an example, consider the use of the identity loss in CycleGAN\citep{zhu2017unpaired}, where it is used to maintain the color composition of the input and output images consistent. This is achieved by minimizing the $L1$ distance between the input modern photo y and the generated photo $G(y)$ (as shown in the upper part of \cref{fig3}). The Identity loss $L_{i d}$ is formulated as follows:

\begin{equation}
    L_{i d}(G, F)=\mathrm{E}_{x \sim p_{\text {data }}(x)}\big[\|F(x)-x \|_1\big]+\mathrm{E}_{y \sim p_{\text {data }}(y)}\big[\|G(y)-y \|_1\big]
\end{equation}

\section{Experiments}\label{sec4}

\subsection{Data preparation}
We experiment with two transfer tasks: Ancient landscape painting to modern photos; landscape photos to sketches. The datasets related to the two tasks are described as follows:\\
\textbf{Ancient Chinese landscape Paintings:}

The ancient Chinese landscape paintings in this study were obtained from a high-quality dataset proposed by Xue\citep{xue2021}, which includes photos collected from four open museum galleries: the Smithsonian Freer Gallery, the Metropolitan Museum of Art, the Princeton University Art Museum, and the Harvard University Art Museum. To improve the quality of the images, we removed some blurry or unattractive photos from the collection. Additionally, since the number of collected images was limited, we used image flipping and cropping to expand our dataset. After preprocessing, we obtained a total of 1940 ancient landscape paintings.

For the collected ancient landscape paintings, some elements such as inscriptions, seals, and poems can affect the generation of style. To address this problem, we applied the corresponding constraint functions introduced in \cref{subsub4.72}.\\
\textbf{Modern Landscape Photos:}

To support our research on modern landscape photos, we collected a dataset of modern landscape photos from the internet. We manually selected and cropped the images to ensure that the dataset included a variety of nature and landscape scenes. After collecting and processing the images, we obtained a total of 1794 modern landscape photos.\\
\textbf{Landscape Sketches:}

To obtain data for our target domain (landscape sketches), we collected sketches from the Internet. We noticed that some sketches have large blank areas, which could hinder the training of the model. Therefore, we manually filtered out these sketches and preprocessed the remaining images. After this process, we obtained a total of 1560 landscape sketches. 

We first scale all image pixel sizes to 512 $\times$ 512. The dataset we collected, some of which are shown in \cref{fig2}. In the training and testing division of the experimental dataset, for the two translation tasks, 800 photos are randomly selected as the training set and 200 photos are randomly selected as the testing set.

\subsection{Training details}

The proposed model was implemented using PyTorch \citep{paszke2019} on a computer equipped with an AMD EPYC 7551P CPU and an NVIDIA GeForce RTX 3090 graphics card with 24GB of video memory. The training process utilized the Adam optimizer \citep{kingma2014} with momentum parameters $\beta_1 = 0.5$ and $\beta_2 = 0.999$, and the initial learning rate was set to 0.0002. The hyperparameters $\lambda_{\text {Dual }}=10$, $\lambda_{\text {GAN }}=1$ and $\lambda_{\text {id }}=5$. The network parameters were initialized using a Gaussian distribution with a mean of 0 and a standard deviation of 0.02. The training was conducted for a total of 200 epochs with a batch size of 1. The proposed model has a total of 32.05 million parameters, contributing to its ability to capture and learn complex style and content mappings.

The training time for the ancient Chinese landscape painting to modern photo task is about 39 hours, and the training time for the ancient Chinese landscape painting to line drawing task is about 37 hours. These training times include 200 epochs of processing time. The experiment was conducted on the Ubuntu 20.04.3 operating system. The original size of the input image was 512 $\times$ 512, and the image size was expanded to 588 $\times$ 588 before input into the model, and then randomly cropped to 512 $\times$ 512.

\subsection{Qualitative results and analysis}

\subsubsection{Landscape painting translation task}

In this section, we perform a qualitative comparison of our proposed model for the task of translating ancient Chinese landscape paintings. This comparison aims to evaluate the performance and effectiveness of our model in capturing the unique style and content of these paintings.

We compare our method with two state-of-the-art neural style transfer methods: Gatys et al.\citep{gatys2016}, AdaIN\citep{huang2017}, and four unpaired image-to-image translation methods: CycleGAN\citep{zhu2017unpaired}, AsymmetryGAN\citep{dou2020}, DSTN\citep{qiao2019}, Wang et al.\citep{wang2022}, DSM\citep{peng2023unsupervised}.

For the two neural style transfer methods, Gatys et al. and AdaIN, their inputs are a content image (Landscape painting) and a style image (one of the collected landscape photos). For the five unpaired image-to-image translation methods, CycleGAN, AsymmetryGAN, DSTN, Wang et al., DSM, we retrained each comparison model using our training set, which consists of 1735 images, including 858 collected ancient landscape paintings and 877 modern landscape photos.

\begin{figure}[ht]
\centering
\includegraphics[width=\textwidth]{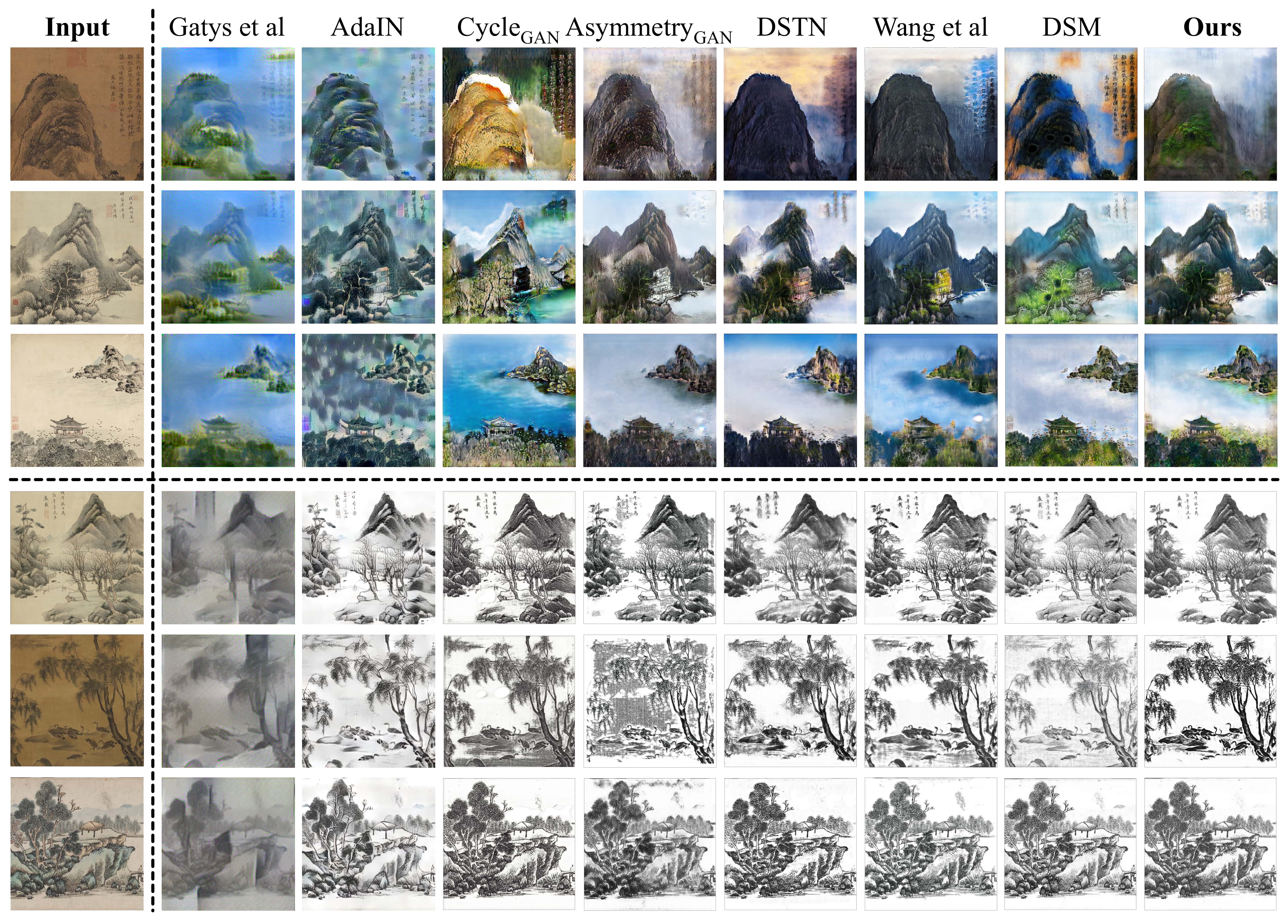}
\caption{Comparison of DLP-GAN with related methods, including neural style transfer based methods Gatys et al.\citep{gatys2016}, AdaIN\citep{huang2017}, and GAN-based image translation methods CycleGAN\citep{zhu2017unpaired}, AsymmetryGAN\citep{dou2020}, DSTN\citep{qiao2019}, Wang et al.\citep{wang2022}, DSM\citep{peng2023unsupervised} on drawing modern landscapes (top), and landscape sketch synthesis (bottom).}
\label{fig7}
\end{figure}

The results of the qualitative comparison are shown in \cref{fig7}. Our method generates higher-quality images that are more similar to real modern photos. The leftmost column shows the input images from ancient landscape paintings. Images generated using the method of Gatys only show low-level texture information, and the textures are not distributed according to the semantic structure. This is particularly evident when generating sketches, where the target domain style is not captured. The images generated by the AdaIN method lack semantic information, and the style is too prominent. When generating modern photos, CycleGAN has a largely random distribution in color space. AsymmetryGAN retain some semantic information and styles in both tasks due to the asymmetrical structure. However, the color space lacks creativity and fails to capture the modern image perfectly. DSTN cannot learn the deep landscape style, and the generated semantics appear distorted with too sharp edges. The method of Wang et al. generates results with artifacts and noise. DSM method provides a richer representation of semantic information in generated sketches. However, it tends to produce sketches with uniform pixel distributions, which may not effectively highlight foreground objects. In contrast, our method generates images that preserve the semantic information well and have natural and realistic color space transitions.

As shown in \cref{fig7}, our GAN-based approach can completely transfer the landscape painting style while preserving more complete semantic details and producing more realistic styles. For example, our approach produces more realistic modern brushstrokes on the edges of mountains and plants. In contrast, the other GAN-based methods shown in the first row fail to capture the semantics of the mountain peaks, and the colors are confusing. In addition, our generated images are more reasonable at different view levels. They have smoother transitions in the sky, river, and mountain edges, which are more consistent with realistic landscape scenery. In generating sketches, our method produces images with more complete semantics and styles, solving the artifacts and noise that existed before. Since sketches lack color information, the original spatial semantic information is more prominent, which is crucial for studying the structural elements of ancient landscape paintings.

\begin{figure}[ht]
\centering
    \begin{minipage}[t]{0.49\textwidth}
        \centering
        \includegraphics[width=1\textwidth]{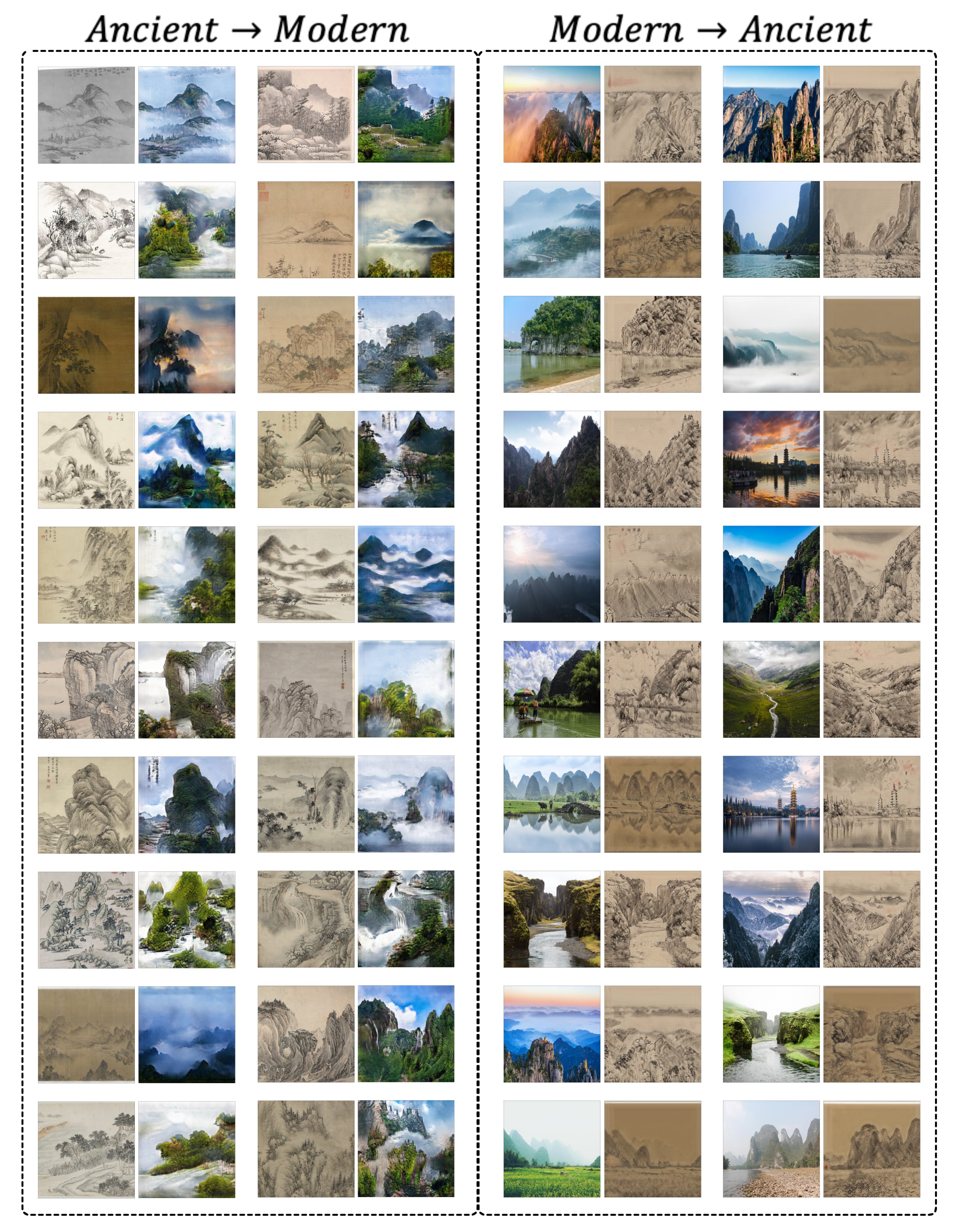}
        \caption{Results of DLP-GAN for drawing landscape evaluated on the test set. Left: translation from ancient landscape painting to modern photos. Right: translation from modern photos to ancient landscape painting.}
        \label{fig8}
    \end{minipage}
    \begin{minipage}[t]{0.49\textwidth}
        \centering
        \includegraphics[width=1\textwidth]{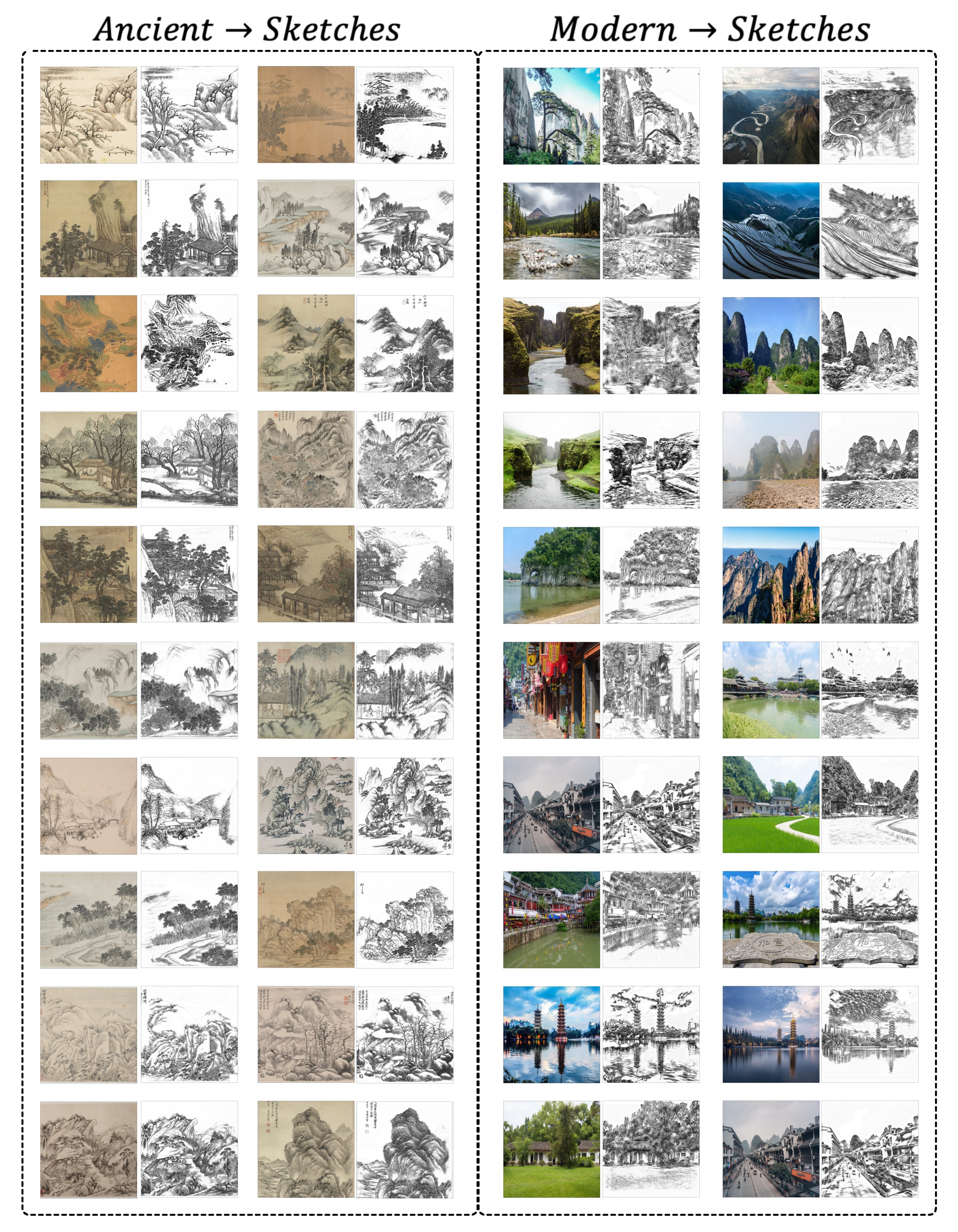}
        \caption{Results of DLP-GAN for drawing landscape sketch evaluated on the test set. Left: translation from ancient landscape painting to sketches. Right: translation from modern photos to sketches.}
        \label{fig9}
        \end{minipage}
\end{figure}

To show more results, we evaluate the performance of our model on two tasks: bidirectional translation and transferability. \Cref{fig8} and \cref{fig9} show the results of these tasks, respectively. In the landscape photo task, we use DLP-GAN to translate images from ancient to modern styles and vice versa. The results shown in \cref{fig8} illustrate that our method can achieve high-quality style transfer. Our approach also extends to sketch synthesis with the same hyperparameter configuration as the landscape photo task. \cref{fig9} shows qualitative results of our model trained on the ancient dataset and applied to the modern dataset. The results show that our method can accurately convey semantic information in both directions.

\begin{figure}[ht] 
\centering
\includegraphics[width=0.9\textwidth]{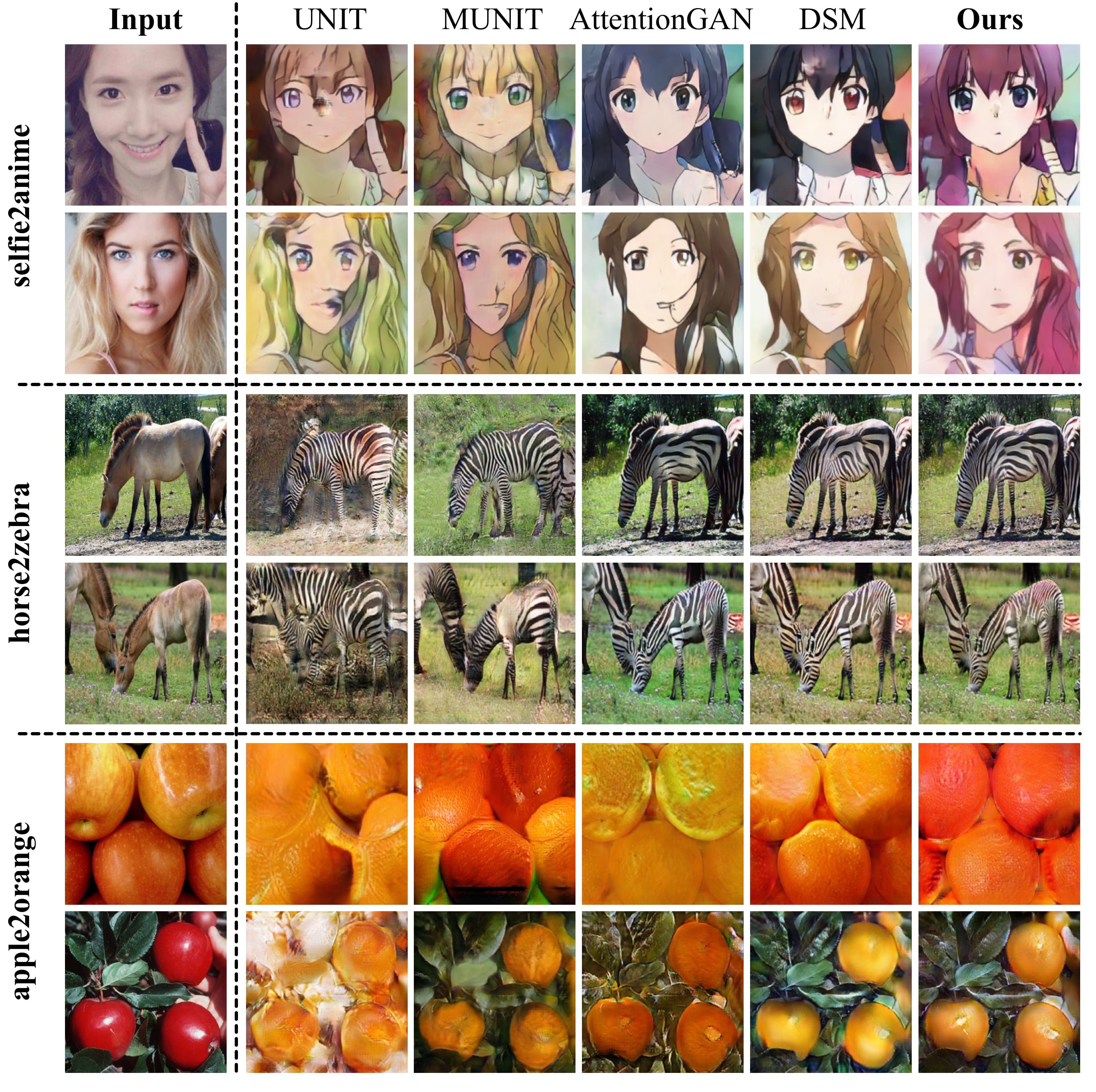}
\caption{Comparison of our proposed method with related methods on three image translation tasks, including four Unpaired image-to-image translation methods: UNIT\cite{liu2017unsupervised}, MUNIT\cite{huang2018multimodal}, AttentionGAN\citep{tang2021attentiongan}, DSM\citep{peng2023unsupervised}.}
\label{fig12}
\end{figure}

\subsubsection{Other translation tasks}

To further evaluate the generalization ability of our model, we conduct qualitative analysis in this section. We compare our model with four state-of-the-art unpaired image-to-image translation methods: UNIT\cite{liu2017unsupervised}, MUNIT\cite{huang2018multimodal}, AttentionGAN\citep{tang2021attentiongan}, DSM\citep{peng2023unsupervised} on several common translation tasks, including selfie2anime, horse2zebra and apple2orange. This analysis provides insight into the strengths and limitations of our model compared to state-of-the-art methods in the field.

A qualitative comparison of the three image-to-image translation tasks is shown in \cref{fig12}. UNIT and MUNIT adopt feature-level cycle-consistency loss, which less constraints the results at the image level, making the generated image deformed. Attention GAN adds attention mask and content mask, which makes the model pay more attention to the foreground of the image, resulting in more changes in the background part of the generated image. DSM, being a state-of-the-art method based on distributional semantics, serves as a benchmark for comparison.

Upon examination, we observe that the overall image generated by our method is slightly better than DSM. Notably, in the third row of the \cref{fig12}, the stripes generated by our method exhibit greater completeness. However, when it comes to capturing fine details, such as the finger region in the first row, our method exhibits slight deviations. This limitation is due to our model's emphasis on the overall translation of landscape paintings, which may lead to slight differences in translating images that contain a single foreground. Regarding this part we will further improve the research in future work.

Through these comparisons and analyses, we aim to demonstrate the effectiveness and generalization of our proposed model in various translation tasks, as well as its potential to outperform existing state-of-the-art methods.

\begin{table}[ht]
\caption{Quantitative comparison with related methods in FID, FID, PSNR and SSIM. Landscape and Sketches denote different transfer tasks respectively. Landscape: ancient landscape to modern photos; Sketches: ancient landscape to sketches. }\label{tab1}
\centering
\resizebox{\textwidth}{!}{
\begin{tabular}{ccccccccc}
\toprule
\multirow{2}{*}{Method}& \multicolumn{2}{c}{FID $\downarrow$} &  \multicolumn{2}{c}{KID $\downarrow$}&  \multicolumn{2}{c}{PSNR $\uparrow$}&  \multicolumn{2}{c}{SSIM $\uparrow$}\\
\cmidrule(r){2-3} \cmidrule(lr){4-5} \cmidrule(lr){6-7} \cmidrule(lr){8-9}
& Landscape& Sketches& Landscape& Sketches& Landscape& Sketches& Landscape& Sketches\\
\midrule
Gatys et al.\citep{gatys2016}& 333.629& 495.327& 0.728& 1.043& 13.013& \textbf{19.412}& 0.459& 0.513\\
AdaIN\citep{huang2017}& 321.893& 274.288& 0.664& 0.578& 13.515& 12.244& 0.408& 0.458\\
CycleGAN\citep{zhu2017unpaired}& 296.965& 270.634& 0.752& 0.582& 10.454& 12.647& 0.497& 0.378\\
AsymmetryGAN\citep{dou2020}& 219.342& 248.027& 0.674& 0.532& \textbf{13.753}& 12.293& 0.515& 0.375\\
DSTN\citep{qiao2019}& 225.294& 209.045& 0.693& 0.448& 12.959& 12.698& 0.591& 0.404\\
Wang et al.\citep{wang2022}& 217.326& 246.921& 0.688& 0.534& 11.684& 11.912& 0.393& 0.415\\
DSM\citep{peng2023unsupervised}& 220.823& 183.384& 0.681& 0.395& 11.973& 12.816& 0.492& 0.542\\
Ours& \textbf{198.835}&\textbf{175.368} &\textbf{0.623} &\textbf{0.386} & 12.328& 12.439& \textbf{0.609}& \textbf{0.552}\\
\bottomrule
\end{tabular}}
\end{table}

\subsection{Quantitative evaluation}

To objectively quantify the results generated by our method, we introduce four evaluation metrics: distribution-based metrics (Frechet Inception Distance (FID) and Kernel Inception Distance (KID)), as well as image-based metrics (Peak Signal-to-Noise Ratio (PSNR) and Structural Similarity (SSIM)). These metrics are described as follows:

\textbf{FID} \citep{heusel2017} measures the similarity of the generated images by calculating the mean and variance of the features obtained from the Inception V3 network. Lower scores indicate greater similarity between the two domains.

\textbf{KID} \citep{binkowski2018} calculates the square of the maximum mean difference between the two sets of image features and represents the distance of feature distribution between the input image and the generated image. A smaller KID value indicates a greater similarity between the two domains. This metric can be used to evaluate the quality of the rendered image to some extent.

\textbf{PSNR and SSIM} \citep{hore2010} are widely used image evaluation metrics to measure the similarity between two images. PSNR is calculated by the error between the input and generated image. SSIM measures the similarity between two images based on their brightness, contrast, and structure. Higher values of PSNR and SSIM indicate better quality of the generated image.

The quantitative comparison results of our experiments are shown in \cref{tab1}. It can be found that our method outperforms the other approaches across all metrics except for PSNR. It is important to note that PSNR primarily focuses on pixel-level similarity between images and does not fully account for other important visual features such as color, texture, and spatial arrangement, which are perceived by the human visual system. \Cref{fig7} illustrates this, especially in the sketching task. Gatys method achieves the highest PSNR score due to its darker background color, resulting in relatively high pixel-level similarity. However, instead of dark backgrounds, our goal in the sketching task is to highlight interesting foreground elements. Despite the variance in PSNR results, our method consistently outperforms other methods on the remaining three metrics, indicating its superior ability to generate high-quality modern photos.

\subsection{User study}
This section provides detailed information about the human perceptual data collection process. We compare our method with the following GAN-based image translation methods: CycleGAN, AsymmetryGAN, DSTN, Wang et al., and DSM. The human perception results are shown in \cref{tab2}, and it is evident that our results are more popular, which indicates that our method achieves better translation results by balancing semantics and abstraction.

\begin{table}[htbp]
\caption{ Results of user studies on the generation effects of different models. The numbers are the performance of votes obtained by each method. Landscape and Sketches denote different transfer tasks respectively. Landscape: ancient landscape to modern photos; Sketches: ancient landscape to sketches.} \label{tab2}
\centering
\resizebox{\textwidth}{!}{
\begin{tabular}{ccccccc}
\toprule
Method&CycleGAN\citep{zhu2017unpaired}& AsymmetryGAN\citep{dou2020}& DSTN\citep{qiao2019}& Wang et al.\citep{wang2022}& DSM\citep{peng2023unsupervised} &Ours\\
\midrule
landscape & 12.86\%& 15.71\%& 15\%& 17.14\%& 16.43\% & \textbf{22.86\%}\\
sketches& 15.71\%& 14.3\%& 17.14\%& 13.57\%& 18.57\% &\textbf{20.71\%}\\
\bottomrule
\end{tabular}}
\end{table}

\begin{figure}[htbp]
\centering
\includegraphics[width=0.7\textwidth]{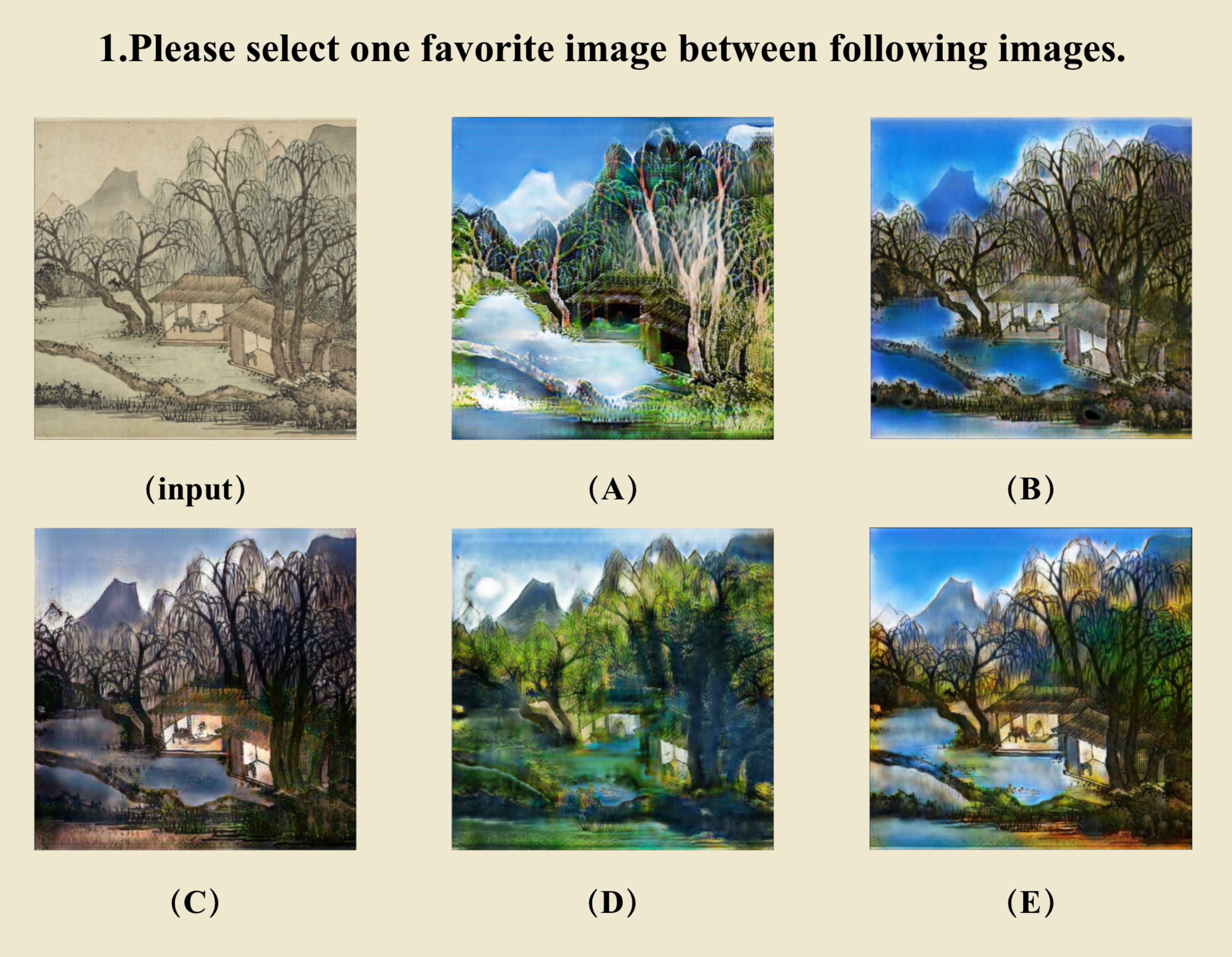}
\caption{The screen shot of user study. Five GAN-based stylized results are shown to the participants at the same time.}
\label{fig10}
\end{figure}

\textbf{Basic setup}: The user study involved 140 participants who received 24 sets of photos randomly selected from the test set. The results for 12 sets are reported in \cref{tab2}, while the results for the remaining 12 sets are shown in \cref{tab3}. The task of the participants is to choose the image that best represents the style of modern realistic painting. To facilitate this task, a dedicated web page (depicted in \cref{fig10}) was utilized, presenting stylized images generated by various methods.

\textbf{Filtering process}: To mitigate potential biases in data collection due to subjective perceptions of painting beauty, we implemented the following measures: (1) Participants with at least a bachelor's degree or equivalent were included in the study. (2) We included repeated groups of images, specifically the second group being identical to the sixth group. This allowed us to assess the consistency of participants' choices before and after. (3) If a participant selected inconsistent results in the second and sixth sets of images, indicating a lack of reliability in their responses, we excluded the data collected from that participant.

\subsection{Ablation study}
In this part, we analyze loss functions and parameters of our model.

\subsubsection{Analysis of loss function}\label{loss4.6.1}
We performed ablation experiments to verify the role of each loss in our method. We trained four model versions on two tasks: only $L_{\text {LSGAN }}$ loss, without $L_{\text {id }}$ loss, without $L_{\text {Semantic }}$ loss, and without $L_{\text {Feature }}$ loss. We then compared each model with our full method using the perceptual study setup described above, and reported the percentage of users who selected our full method in each ablation. As shown in \cref{tab3}, the largest number of people chose our full model, indicating the effectiveness of our method.

\begin{figure}[htb]
\centering
\includegraphics[width=0.85\textwidth]{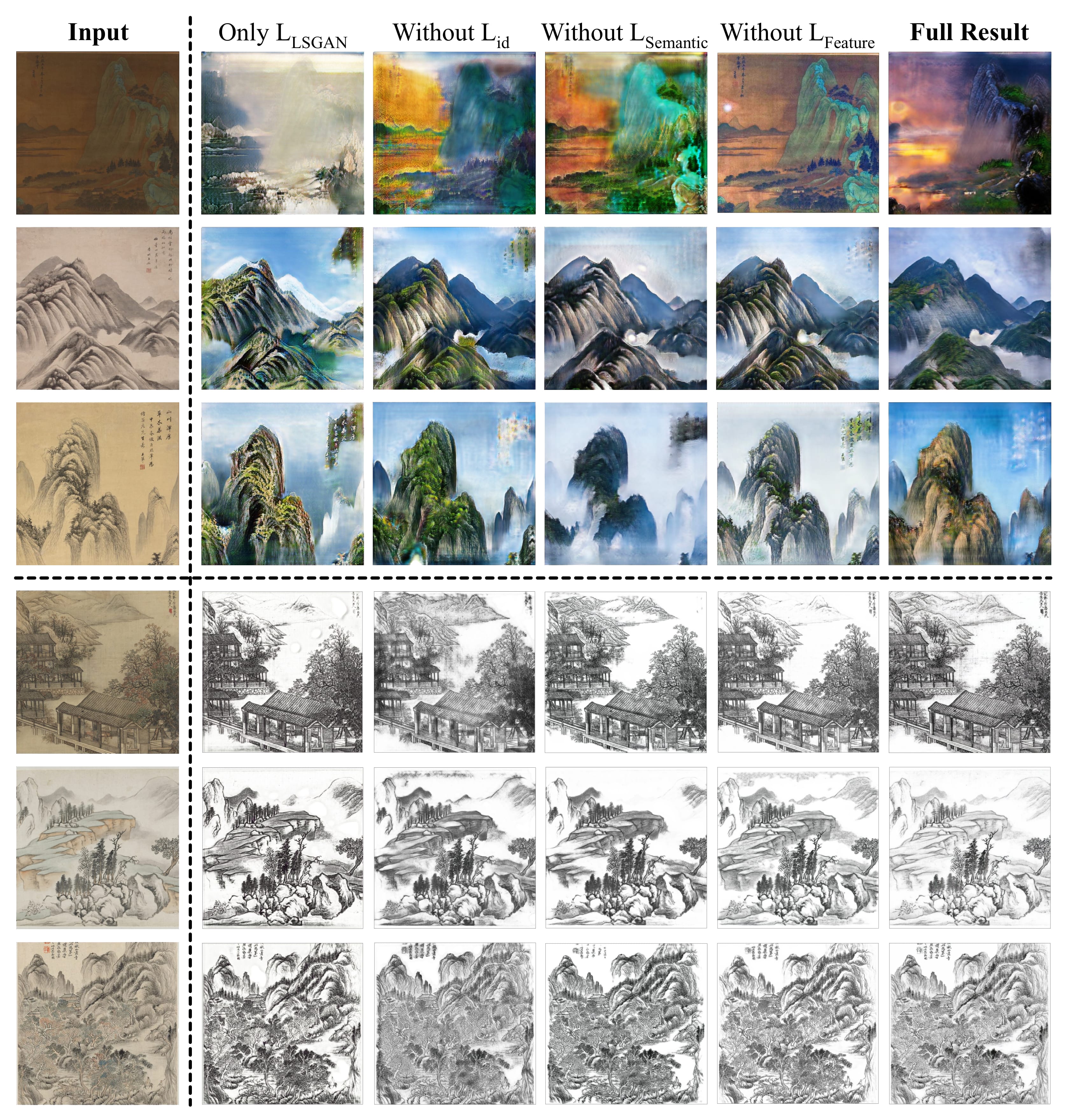}
\caption{Ablation study of the proposed LSGAN loss $L_{\text {LSGAN }}$, identity loss $L_{\text {id }}$, semantic loss $L_{\text {Semantic }}$ and feature loss $L_{\text {Feature }}$ for two tasks. Using all losses yields the most attractive results. }
\label{fig11}
\end{figure}

\begin{table}[htbp]
\caption{Results of user studies on loss function analysis in \cref{loss4.6.1}.}\label{tab3}
\centering
\resizebox{\textwidth}{!}{
\begin{tabular}{cccccc}
\toprule
Model version&only $L_{\text {LSGAN }}$& no $L_{\text {id }}$ & no $L_{\text {Semantic }}$& no $L_{\text {Feature }}$& Full Result\\
\midrule
landscape & 21.90\%& 17.14\%& 14.76\%& 12.38\%& \textbf{33.81\%}\\
sketches& 16.19\%& 13.33\%& 21.43\%& 13.33\%& \textbf{35.71\%}\\
\bottomrule
\end{tabular}}
\end{table}

As shown in \cref{fig11}. Training results using only $L_{\text {LSGAN }}$ are affected by the absence and deformation of the content, which can limit the results to the outline of the painting but lack recognition. When the identity loss $L_{\text {id }}$ is removed, the result is uncoordinated colors and a lack of layers. Removing the semantic loss $L_{\text {Semantic }}$ leads to more abstract and stylized landscape paintings. On sketches, the effect of improving stylization intensity is less obvious, possibly due to the unimodal style distribution of sketches and the ability of the feature loss to fully capture the target domain style. Removing the feature loss $L_{\text {Feature }}$ leads to strengthened semantics and clearer edges, but does not produce a sense of sharpness during actual drawing. The images generated by these methods have unexpected noise and artifacts, which are not conducive to studying the structure of the original domain images or generating high-quality photos. Our complete approach to drawing photos achieves an optimal balance between stylization and abstraction. For example, in the first row, the sunset and light on the water are realistic, and the mountain part is more layered. In sketches, integrity is improved, closer to how we would draw.

\begin{table}[htbp]
\caption{Quantitative experimental results of FID and KID about loss function on sketch tasks.}\label{tab4}
\centering
\resizebox{\textwidth}{!}{
\begin{tabular}{cccccc}
\toprule
Model version&only $L_{\text {LSGAN }}$& no $L_{\text {id }}$ & no $L_{\text {Semantic }}$& no $L_{\text {Feature }}$& full method\\
\midrule
FID$\downarrow$& 270.634& 194.138& 201.143& 208.192& \textbf{175.368}\\
KID$\downarrow$& 0.582& 0.405& 0.434& 0.437& \textbf{0.386}\\
\bottomrule
\end{tabular}}
\end{table}

We present the results of the quantitative experimental comparison of the loss function in \cref{tab4}. We chose to focus on the sketch task because the blank background emphasizes the noise and artifacts, leading to a more significant quantitative effect. In contrast, we study the landscape painting task in the parameter ablation experiment.

These results show that both semantic-consistency loss and feature-consistency loss help to preserve spatial information and are complementary to each other: (i) the feature loss works in a more global and general way, it adds to the abstraction of the scene and helps to preserve contours (as lines are more easily missing in mountain peaks, plants and river regions, where the effects are more noticeable), and (ii) as a comparison, the semantic loss works in a local way, dedicated to image stroke edges, improving drawings and eliminating artifacts in these local regions.

\subsubsection{Influence of weighting parameters}\label{subsub4.72}

\begin{figure}[htb]
\centering
\includegraphics[width=0.8\textwidth]{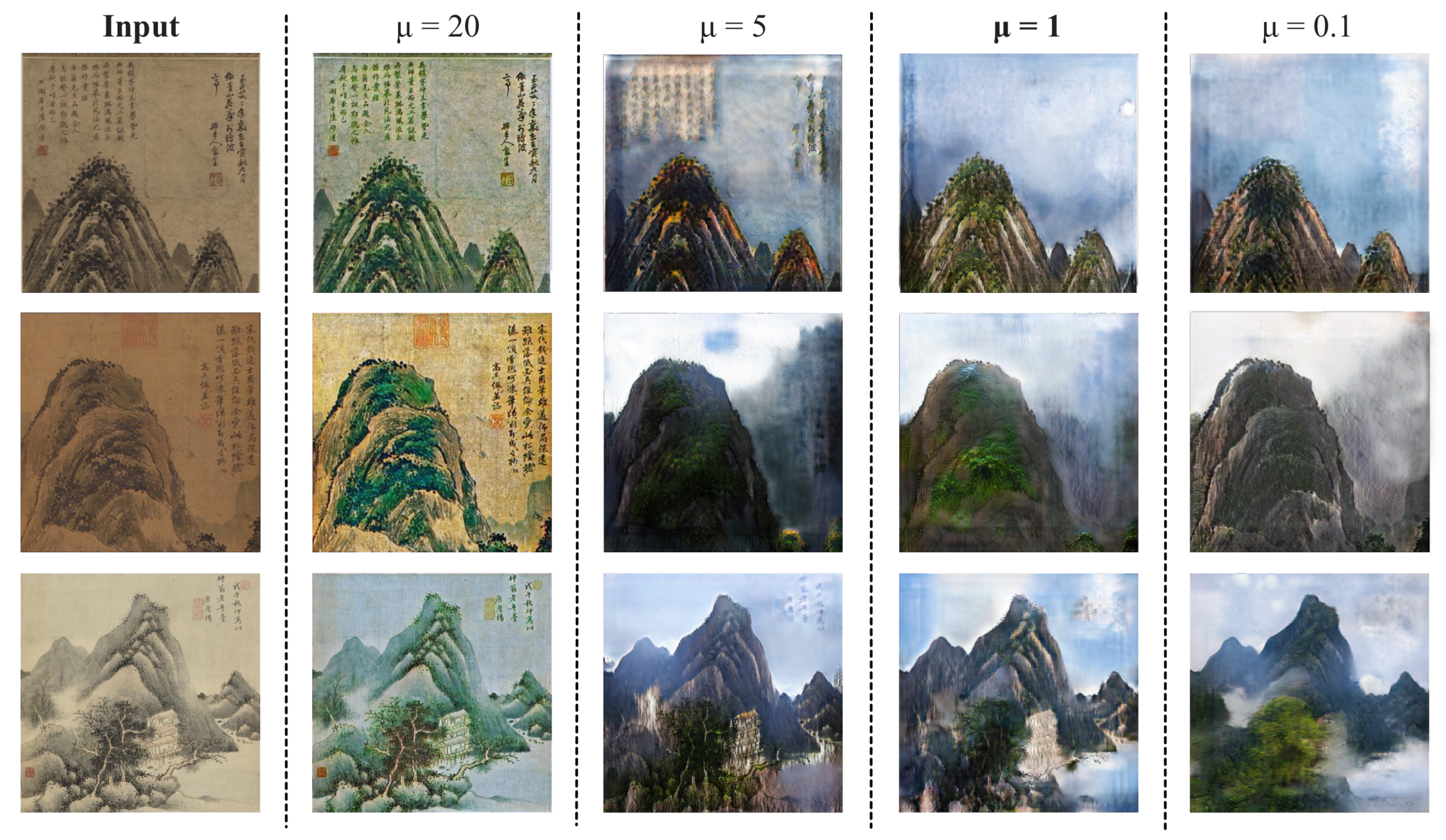}
\caption{Ablation study of the proposed parameter $\mu$ from $L_{\text {Dual }} $ loss, $\mu = \{ 20, 5, 1, 0.1\}.$ }
\label{fig6}
\end{figure}

This section demonstrates the results of the hyperparametric ablation study. As mentioned above, previous methods manually removed these elements in the pre-processing stage for the collected ancient landscape paintings because of elements that will affect the result, such as verse and seal. Instead, we explore a novel end-to-end approach. The method results are shown in \cref{fig6}. To mask these elements, our framework controls the weight between $L_{\text {Semantic }}$ and $L_{\text {Feature }}$ by adjusting the value of $\mu$ in dual-consistency loss. From the above Eq.(\ref{equ9}), as the value of $\mu$ increases, the weight of semantic loss becomes larger, which develops the ability of the model to retain the content. Conversely, the model tends to capture the image style. To explore the influence of trade-off parameter, we do experiments by setting $\mu$ = \{ 20, 5, 1, 0.1\}.

\begin{table}[htbp]
\caption{Quantitative evaluation was conducted on the landscape painting task in FID and KID for parameters $\mu$ = \{ 20, 5, 1, 0.1\}.  }\label{tab5}
\centering
\begin{tabular}{ccccc}
\toprule
Metrics &$\mu$ = 20 & $\mu$ = 5 & $\mu$ = 1 (Ours) & $\mu$ = 0.1\\
\midrule
FID$\downarrow$& \textbf{188.193}& 234.313& 198.835& 253.831\\
KID$\downarrow$& \textbf{0.549}& 0.682& 0.623& 0.719\\
\bottomrule
\end{tabular}
\end{table}
As shown by \cref{fig6} and \cref{tab5}, when $\mu$ = 20, the model has the greatest ability to retain content, the details on the picture are well-preserved. The background color tends to be uniform (mountains, plants, and verses have the same color). Although FID and KID are the lowest, the pictures lack abstraction and cannot express human creativity. When $\mu$ = 0.1, the image translation result has a weak content consistency, and the poems are fully masked, which brings another disadvantage in that the color of the image becomes unrealistic. For example, the whole image is white, and the edge of the brush stroke is more blurred. With the $\mu$ value decreasing, the style is enhanced, and when $\mu$ = 1, the content and style reach the optimal balance, while the value of the quantitative comparison is also relatively minimal, so we adopt this parameter setting.

\section{Conclusion}\label{sec5}

In this paper, we propose a method called DLP-GAN, which is based on asymmetric cycle-consistent adversarial architectures, to automatically draw landscape photos while achieving semantic-based style transfer between unpaired datasets. The architecture includes a generator based on a dense-fusion module, which can capture more input elements to enhance feature propagation. At the same time, a dual-consistency loss is introduced to improve the realism of the generated photos. Experiments demonstrate that our model can generate landscape photos that better trade-off between artistic abstraction and realism. We believe this work can help us to research the style transfer task of landscape painting from another perspective.

As for our future work, it is a valuable topic to explore more types of paintings to generate modern photos, meanwhile further improving the quality of the generated images.

\backmatter





\bibliography{Manuscript}


\begin{thebibliography}{49}
\ifx \bisbn   \undefined \def \bisbn  #1{ISBN #1}\fi
\ifx \binits  \undefined \def \binits#1{#1}\fi
\ifx \bauthor  \undefined \def \bauthor#1{#1}\fi
\ifx \batitle  \undefined \def \batitle#1{#1}\fi
\ifx \bjtitle  \undefined \def \bjtitle#1{#1}\fi
\ifx \bvolume  \undefined \def \bvolume#1{\textbf{#1}}\fi
\ifx \byear  \undefined \def \byear#1{#1}\fi
\ifx \bissue  \undefined \def \bissue#1{#1}\fi
\ifx \bfpage  \undefined \def \bfpage#1{#1}\fi
\ifx \blpage  \undefined \def \blpage #1{#1}\fi
\ifx \burl  \undefined \def \burl#1{\textsf{#1}}\fi
\ifx \doiurl  \undefined \def \doiurl#1{\url{https://doi.org/#1}}\fi
\ifx \betal  \undefined \def \betal{\textit{et al.}}\fi
\ifx \binstitute  \undefined \def \binstitute#1{#1}\fi
\ifx \binstitutionaled  \undefined \def \binstitutionaled#1{#1}\fi
\ifx \bctitle  \undefined \def \bctitle#1{#1}\fi
\ifx \beditor  \undefined \def \beditor#1{#1}\fi
\ifx \bpublisher  \undefined \def \bpublisher#1{#1}\fi
\ifx \bbtitle  \undefined \def \bbtitle#1{#1}\fi
\ifx \bedition  \undefined \def \bedition#1{#1}\fi
\ifx \bseriesno  \undefined \def \bseriesno#1{#1}\fi
\ifx \blocation  \undefined \def \blocation#1{#1}\fi
\ifx \bsertitle  \undefined \def \bsertitle#1{#1}\fi
\ifx \bsnm \undefined \def \bsnm#1{#1}\fi
\ifx \bsuffix \undefined \def \bsuffix#1{#1}\fi
\ifx \bparticle \undefined \def \bparticle#1{#1}\fi
\ifx \barticle \undefined \def \barticle#1{#1}\fi
\bibcommenthead
\ifx \bconfdate \undefined \def \bconfdate #1{#1}\fi
\ifx \botherref \undefined \def \botherref #1{#1}\fi
\ifx \url \undefined \def \url#1{\textsf{#1}}\fi
\ifx \bchapter \undefined \def \bchapter#1{#1}\fi
\ifx \bbook \undefined \def \bbook#1{#1}\fi
\ifx \bcomment \undefined \def \bcomment#1{#1}\fi
\ifx \oauthor \undefined \def \oauthor#1{#1}\fi
\ifx \citeauthoryear \undefined \def \citeauthoryear#1{#1}\fi
\ifx \endbibitem  \undefined \def \endbibitem {}\fi
\ifx \bconflocation  \undefined \def \bconflocation#1{#1}\fi
\ifx \arxivurl  \undefined \def \arxivurl#1{\textsf{#1}}\fi
\csname PreBibitemsHook\endcsname

\bibitem{liu2021basic}
\begin{bchapter}
\bauthor{\bsnm{Liu}, \binits{L.}}:
\bctitle{The basic features of traditional chinese landscape painting}.
In: \bbtitle{The 5th International Conference on Art Studies: Research, Experience, Education (ICASSEE 2021)},
vol. \bseriesno{1},
pp. \bfpage{17}--\blpage{27}
(\byear{2021}).
\doiurl{10.5117/9789048557240/ICASSEE.2021.003}.
\bcomment{Amsterdam University Press}
\end{bchapter}
\endbibitem

\bibitem{li2017}
\begin{botherref}
\oauthor{\bsnm{Li}, \binits{Y.}},
\oauthor{\bsnm{Fang}, \binits{C.}},
\oauthor{\bsnm{Yang}, \binits{J.}},
\oauthor{\bsnm{Wang}, \binits{Z.}},
\oauthor{\bsnm{Lu}, \binits{X.}},
\oauthor{\bsnm{Yang}, \binits{M.-H.}}:
Universal style transfer via feature transforms.
Advances in neural information processing systems
\textbf{30}
(2017)
\end{botherref}
\endbibitem

\bibitem{gatys2016}
\begin{bchapter}
\bauthor{\bsnm{Gatys}, \binits{L.A.}},
\bauthor{\bsnm{Ecker}, \binits{A.S.}},
\bauthor{\bsnm{Bethge}, \binits{M.}}:
\bctitle{Image style transfer using convolutional neural networks}.
In: \bbtitle{Proceedings of the IEEE Conference on Computer Vision and Pattern Recognition},
pp. \bfpage{2414}--\blpage{2423}
(\byear{2016}).
\doiurl{10.1109/cvpr.2016.265}
\end{bchapter}
\endbibitem

\bibitem{johnson2016}
\begin{bchapter}
\bauthor{\bsnm{Johnson}, \binits{J.}},
\bauthor{\bsnm{Alahi}, \binits{A.}},
\bauthor{\bsnm{Fei-Fei}, \binits{L.}}:
\bctitle{Perceptual losses for real-time style transfer and super-resolution}.
In: \bbtitle{European Conference on Computer Vision},
pp. \bfpage{694}--\blpage{711}
(\byear{2016}).
\doiurl{10.1007/978-3-319-46475-6_43}.
\bcomment{Springer}
\end{bchapter}
\endbibitem

\bibitem{zhu2017unpaired}
\begin{bchapter}
\bauthor{\bsnm{Zhu}, \binits{J.-Y.}},
\bauthor{\bsnm{Park}, \binits{T.}},
\bauthor{\bsnm{Isola}, \binits{P.}},
\bauthor{\bsnm{Efros}, \binits{A.A.}}:
\bctitle{Unpaired image-to-image translation using cycle-consistent adversarial networks}.
In: \bbtitle{Proceedings of the IEEE International Conference on Computer Vision},
pp. \bfpage{2223}--\blpage{2232}
(\byear{2017}).
\doiurl{10.1109/iccv.2017.244}
\end{bchapter}
\endbibitem

\bibitem{zhu2017toward}
\begin{botherref}
\oauthor{\bsnm{Zhu}, \binits{J.-Y.}},
\oauthor{\bsnm{Zhang}, \binits{R.}},
\oauthor{\bsnm{Pathak}, \binits{D.}},
\oauthor{\bsnm{Darrell}, \binits{T.}},
\oauthor{\bsnm{Efros}, \binits{A.A.}},
\oauthor{\bsnm{Wang}, \binits{O.}},
\oauthor{\bsnm{Shechtman}, \binits{E.}}:
Toward multimodal image-to-image translation.
Advances in neural information processing systems
\textbf{30}
(2017)
\end{botherref}
\endbibitem

\bibitem{isola2017}
\begin{bchapter}
\bauthor{\bsnm{Isola}, \binits{P.}},
\bauthor{\bsnm{Zhu}, \binits{J.-Y.}},
\bauthor{\bsnm{Zhou}, \binits{T.}},
\bauthor{\bsnm{Efros}, \binits{A.A.}}:
\bctitle{Image-to-image translation with conditional adversarial networks}.
In: \bbtitle{Proceedings of the IEEE Conference on Computer Vision and Pattern Recognition},
pp. \bfpage{1125}--\blpage{1134}
(\byear{2017}).
\doiurl{10.1109/cvpr.2017.632}
\end{bchapter}
\endbibitem

\bibitem{li2020sdp}
\begin{barticle}
\bauthor{\bsnm{Li}, \binits{R.}},
\bauthor{\bsnm{Wu}, \binits{C.-H.}},
\bauthor{\bsnm{Liu}, \binits{S.}},
\bauthor{\bsnm{Wang}, \binits{J.}},
\bauthor{\bsnm{Wang}, \binits{G.}},
\bauthor{\bsnm{Liu}, \binits{G.}},
\bauthor{\bsnm{Zeng}, \binits{B.}}:
\batitle{Sdp-gan: saliency detail preservation generative adversarial networks for high perceptual quality style transfer}.
\bjtitle{IEEE Transactions on Image Processing}
\bvolume{30},
\bfpage{374}--\blpage{385}
(\byear{2020}).
\doiurl{10.1109/TIP.2020.3036754}
\end{barticle}
\endbibitem

\bibitem{lin2021drafting}
\begin{bchapter}
\bauthor{\bsnm{Lin}, \binits{T.}},
\bauthor{\bsnm{Ma}, \binits{Z.}},
\bauthor{\bsnm{Li}, \binits{F.}},
\bauthor{\bsnm{He}, \binits{D.}},
\bauthor{\bsnm{Li}, \binits{X.}},
\bauthor{\bsnm{Ding}, \binits{E.}},
\bauthor{\bsnm{Wang}, \binits{N.}},
\bauthor{\bsnm{Li}, \binits{J.}},
\bauthor{\bsnm{Gao}, \binits{X.}}:
\bctitle{Drafting and revision: Laplacian pyramid network for fast high-quality artistic style transfer}.
In: \bbtitle{Proceedings of the IEEE/CVF Conference on Computer Vision and Pattern Recognition},
pp. \bfpage{5141}--\blpage{5150}
(\byear{2021}).
\doiurl{10.1109/cvpr46437.2021.00510}
\end{bchapter}
\endbibitem

\bibitem{liu2021adaattn}
\begin{bchapter}
\bauthor{\bsnm{Liu}, \binits{S.}},
\bauthor{\bsnm{Lin}, \binits{T.}},
\bauthor{\bsnm{He}, \binits{D.}},
\bauthor{\bsnm{Li}, \binits{F.}},
\bauthor{\bsnm{Wang}, \binits{M.}},
\bauthor{\bsnm{Li}, \binits{X.}},
\bauthor{\bsnm{Sun}, \binits{Z.}},
\bauthor{\bsnm{Li}, \binits{Q.}},
\bauthor{\bsnm{Ding}, \binits{E.}}:
\bctitle{Adaattn: Revisit attention mechanism in arbitrary neural style transfer}.
In: \bbtitle{Proceedings of the IEEE/CVF International Conference on Computer Vision},
pp. \bfpage{6649}--\blpage{6658}
(\byear{2021}).
\doiurl{10.1109/iccv48922.2021.00658}
\end{bchapter}
\endbibitem

\bibitem{peng2022contour}
\begin{botherref}
\oauthor{\bsnm{Peng}, \binits{X.}},
\oauthor{\bsnm{Peng}, \binits{S.}},
\oauthor{\bsnm{Hu}, \binits{Q.}},
\oauthor{\bsnm{Peng}, \binits{J.}},
\oauthor{\bsnm{Wang}, \binits{J.}},
\oauthor{\bsnm{Liu}, \binits{X.}},
\oauthor{\bsnm{Fan}, \binits{J.}}:
Contour-enhanced cyclegan framework for style transfer from scenery photos to chinese landscape paintings.
Neural Computing and Applications,
1--22
(2022).
\doiurl{10.1007/s00521-022-07432-w}
\end{botherref}
\endbibitem

\bibitem{zheng2018}
\begin{bchapter}
\bauthor{\bsnm{Zheng}, \binits{C.}},
\bauthor{\bsnm{Zhang}, \binits{Y.}}:
\bctitle{Two-stage color ink painting style transfer via convolution neural network}.
In: \bbtitle{2018 15th International Symposium on Pervasive Systems, Algorithms and Networks (I-SPAN)},
pp. \bfpage{193}--\blpage{200}
(\byear{2018}).
\doiurl{10.1109/i-span.2018.00039}.
\bcomment{IEEE}
\end{bchapter}
\endbibitem

\bibitem{zhou2019}
\begin{bchapter}
\bauthor{\bsnm{Zhou}, \binits{L.}},
\bauthor{\bsnm{Wang}, \binits{Q.-F.}},
\bauthor{\bsnm{Huang}, \binits{K.}},
\bauthor{\bsnm{Lo}, \binits{C.-H.}}:
\bctitle{An interactive and generative approach for chinese shanshui painting document}.
In: \bbtitle{2019 International Conference on Document Analysis and Recognition (ICDAR)},
pp. \bfpage{819}--\blpage{824}
(\byear{2019}).
\doiurl{10.1109/icdar.2019.00136}.
\bcomment{IEEE}
\end{bchapter}
\endbibitem

\bibitem{goodfellow2014}
\begin{barticle}
\bauthor{\bsnm{Goodfellow}, \binits{I.}},
\bauthor{\bsnm{Pouget-Abadie}, \binits{J.}},
\bauthor{\bsnm{Mirza}, \binits{M.}},
\bauthor{\bsnm{Xu}, \binits{B.}},
\bauthor{\bsnm{Warde-Farley}, \binits{D.}},
\bauthor{\bsnm{Ozair}, \binits{S.}},
\bauthor{\bsnm{Courville}, \binits{A.}},
\bauthor{\bsnm{Bengio}, \binits{Y.}}:
\batitle{Generative adversarial networks}.
\bjtitle{Communications of the ACM}
\bvolume{63}(\bissue{11}),
\bfpage{139}--\blpage{144}
(\byear{2020}).
\doiurl{10.1145/3422622}
\end{barticle}
\endbibitem

\bibitem{bharti2022emocgan}
\begin{barticle}
\bauthor{\bsnm{Bharti}, \binits{V.}},
\bauthor{\bsnm{Biswas}, \binits{B.}},
\bauthor{\bsnm{Shukla}, \binits{K.K.}}:
\batitle{Emocgan: a novel evolutionary multiobjective cyclic generative adversarial network and its application to unpaired image translation}.
\bjtitle{Neural Computing and Applications}
\bvolume{34}(\bissue{24}),
\bfpage{21433}--\blpage{21447}
(\byear{2022}).
\doiurl{10.1007/s00521-021-05975-y}
\end{barticle}
\endbibitem

\bibitem{he2018}
\begin{bchapter}
\bauthor{\bsnm{He}, \binits{B.}},
\bauthor{\bsnm{Gao}, \binits{F.}},
\bauthor{\bsnm{Ma}, \binits{D.}},
\bauthor{\bsnm{Shi}, \binits{B.}},
\bauthor{\bsnm{Duan}, \binits{L.-Y.}}:
\bctitle{Chipgan: A generative adversarial network for chinese ink wash painting style transfer}.
In: \bbtitle{Proceedings of the 26th ACM International Conference on Multimedia},
pp. \bfpage{1172}--\blpage{1180}
(\byear{2018}).
\doiurl{10.1145/3240508.3240655}
\end{bchapter}
\endbibitem

\bibitem{wang2022}
\begin{barticle}
\bauthor{\bsnm{Wang}, \binits{W.}},
\bauthor{\bsnm{Li}, \binits{Y.}},
\bauthor{\bsnm{Ye}, \binits{H.}},
\bauthor{\bsnm{Ye}, \binits{F.}},
\bauthor{\bsnm{Xu}, \binits{X.}}:
\batitle{Ink painting style transfer using asymmetric cycle-consistent gan}.
\bjtitle{Available at SSRN 4109972}
(\byear{2022}).
\doiurl{10.2139/ssrn.4109972}
\end{barticle}
\endbibitem

\bibitem{li2018neural}
\begin{bchapter}
\bauthor{\bsnm{Li}, \binits{B.}},
\bauthor{\bsnm{Xiong}, \binits{C.}},
\bauthor{\bsnm{Wu}, \binits{T.}},
\bauthor{\bsnm{Zhou}, \binits{Y.}},
\bauthor{\bsnm{Zhang}, \binits{L.}},
\bauthor{\bsnm{Chu}, \binits{R.}}:
\bctitle{Neural abstract style transfer for chinese traditional painting}.
In: \bbtitle{Asian Conference on Computer Vision},
pp. \bfpage{212}--\blpage{227}
(\byear{2018}).
\doiurl{10.1007/978-3-030-20890-5_14}.
\bcomment{Springer}
\end{bchapter}
\endbibitem

\bibitem{qiao2019}
\begin{bchapter}
\bauthor{\bsnm{Qiao}, \binits{T.}},
\bauthor{\bsnm{Zhang}, \binits{W.}},
\bauthor{\bsnm{Zhang}, \binits{M.}},
\bauthor{\bsnm{Ma}, \binits{Z.}},
\bauthor{\bsnm{Xu}, \binits{D.}}:
\bctitle{Ancient painting to natural image: A new solution for painting processing}.
In: \bbtitle{2019 IEEE Winter Conference on Applications of Computer Vision (WACV)},
pp. \bfpage{521}--\blpage{530}
(\byear{2019}).
\doiurl{10.1109/wacv.2019.00061}
\end{bchapter}
\endbibitem

\bibitem{qin2022towards}
\begin{barticle}
\bauthor{\bsnm{Qin}, \binits{S.}},
\bauthor{\bsnm{Liu}, \binits{S.}}:
\batitle{Towards end-to-end car license plate location and recognition in unconstrained scenarios}.
\bjtitle{Neural Computing and Applications}
\bvolume{34}(\bissue{24}),
\bfpage{21551}--\blpage{21566}
(\byear{2022}).
\doiurl{10.1007/s00521-021-06147-8}
\end{barticle}
\endbibitem

\bibitem{sun2022}
\begin{bchapter}
\bauthor{\bsnm{Sun}, \binits{H.}},
\bauthor{\bsnm{Wu}, \binits{L.}},
\bauthor{\bsnm{Li}, \binits{X.}},
\bauthor{\bsnm{Meng}, \binits{X.}}:
\bctitle{Style-woven attention network for zero-shot ink wash painting style transfer}.
In: \bbtitle{Proceedings of the 2022 International Conference on Multimedia Retrieval},
pp. \bfpage{277}--\blpage{285}
(\byear{2022}).
\doiurl{10.1145/3512527.3531391}
\end{bchapter}
\endbibitem

\bibitem{li2021}
\begin{bchapter}
\bauthor{\bsnm{Li}, \binits{J.}},
\bauthor{\bsnm{Wang}, \binits{Q.}},
\bauthor{\bsnm{Li}, \binits{S.}},
\bauthor{\bsnm{Zhong}, \binits{Q.}},
\bauthor{\bsnm{Zhou}, \binits{Q.}}:
\bctitle{Immersive traditional chinese portrait painting: Research on style transfer and face replacement}.
In: \bbtitle{Chinese Conference on Pattern Recognition and Computer Vision (PRCV)},
pp. \bfpage{192}--\blpage{203}
(\byear{2021}).
\doiurl{10.1007/978-3-030-88007-1_16}.
\bcomment{Springer}
\end{bchapter}
\endbibitem

\bibitem{xue2021}
\begin{bchapter}
\bauthor{\bsnm{Xue}, \binits{A.}}:
\bctitle{End-to-end chinese landscape painting creation using generative adversarial networks}.
In: \bbtitle{Proceedings of the IEEE/CVF Winter Conference on Applications of Computer Vision},
pp. \bfpage{3863}--\blpage{3871}
(\byear{2021}).
\doiurl{10.1109/wacv48630.2021.00391}
\end{bchapter}
\endbibitem

\bibitem{dhariwal2021diffusion}
\begin{barticle}
\bauthor{\bsnm{Dhariwal}, \binits{P.}},
\bauthor{\bsnm{Nichol}, \binits{A.}}:
\batitle{Diffusion models beat gans on image synthesis}.
\bjtitle{Advances in Neural Information Processing Systems}
\bvolume{34},
\bfpage{8780}--\blpage{8794}
(\byear{2021})
\end{barticle}
\endbibitem

\bibitem{ho2020denoising}
\begin{barticle}
\bauthor{\bsnm{Ho}, \binits{J.}},
\bauthor{\bsnm{Jain}, \binits{A.}},
\bauthor{\bsnm{Abbeel}, \binits{P.}}:
\batitle{Denoising diffusion probabilistic models}.
\bjtitle{Advances in Neural Information Processing Systems}
\bvolume{33},
\bfpage{6840}--\blpage{6851}
(\byear{2020})
\end{barticle}
\endbibitem

\bibitem{saharia2022palette}
\begin{bchapter}
\bauthor{\bsnm{Saharia}, \binits{C.}},
\bauthor{\bsnm{Chan}, \binits{W.}},
\bauthor{\bsnm{Chang}, \binits{H.}},
\bauthor{\bsnm{Lee}, \binits{C.}},
\bauthor{\bsnm{Ho}, \binits{J.}},
\bauthor{\bsnm{Salimans}, \binits{T.}},
\bauthor{\bsnm{Fleet}, \binits{D.}},
\bauthor{\bsnm{Norouzi}, \binits{M.}}:
\bctitle{Palette: Image-to-image diffusion models}.
In: \bbtitle{ACM SIGGRAPH 2022 Conference Proceedings},
pp. \bfpage{1}--\blpage{10}
(\byear{2022}).
\doiurl{10.1145/3528233.3530757}
\end{bchapter}
\endbibitem

\bibitem{su2022dual}
\begin{barticle}
\bauthor{\bsnm{Su}, \binits{X.}},
\bauthor{\bsnm{Song}, \binits{J.}},
\bauthor{\bsnm{Meng}, \binits{C.}},
\bauthor{\bsnm{Ermon}, \binits{S.}}:
\batitle{Dual diffusion implicit bridges for image-to-image translation}.
\bjtitle{arXiv preprint arXiv:2203.08382}
(\byear{2022}).
\doiurl{10.48550/arXiv.2203.08382}
\end{barticle}
\endbibitem

\bibitem{li2023bbdm}
\begin{bchapter}
\bauthor{\bsnm{Li}, \binits{B.}},
\bauthor{\bsnm{Xue}, \binits{K.}},
\bauthor{\bsnm{Liu}, \binits{B.}},
\bauthor{\bsnm{Lai}, \binits{Y.-K.}}:
\bctitle{Bbdm: Image-to-image translation with brownian bridge diffusion models}.
In: \bbtitle{Proceedings of the IEEE/CVF Conference on Computer Vision and Pattern Recognition},
pp. \bfpage{1952}--\blpage{1961}
(\byear{2023})
\end{bchapter}
\endbibitem

\bibitem{li2018dense}
\begin{barticle}
\bauthor{\bsnm{Li}, \binits{H.}},
\bauthor{\bsnm{Wu}, \binits{X.-J.}}:
\batitle{Densefuse: A fusion approach to infrared and visible images}.
\bjtitle{IEEE Transactions on Image Processing}
\bvolume{28}(\bissue{5}),
\bfpage{2614}--\blpage{2623}
(\byear{2018}).
\doiurl{10.1109/tip.2018.2887342}
\end{barticle}
\endbibitem

\bibitem{wang2018}
\begin{bchapter}
\bauthor{\bsnm{Wang}, \binits{T.-C.}},
\bauthor{\bsnm{Liu}, \binits{M.-Y.}},
\bauthor{\bsnm{Zhu}, \binits{J.-Y.}},
\bauthor{\bsnm{Tao}, \binits{A.}},
\bauthor{\bsnm{Kautz}, \binits{J.}},
\bauthor{\bsnm{Catanzaro}, \binits{B.}}:
\bctitle{High-resolution image synthesis and semantic manipulation with conditional gans}.
In: \bbtitle{Proceedings of the IEEE Conference on Computer Vision and Pattern Recognition},
pp. \bfpage{8798}--\blpage{8807}
(\byear{2018}).
\doiurl{10.1109/cvpr.2018.00917}
\end{bchapter}
\endbibitem

\bibitem{huang2018multimodal}
\begin{bchapter}
\bauthor{\bsnm{Huang}, \binits{X.}},
\bauthor{\bsnm{Liu}, \binits{M.-Y.}},
\bauthor{\bsnm{Belongie}, \binits{S.}},
\bauthor{\bsnm{Kautz}, \binits{J.}}:
\bctitle{Multimodal unsupervised image-to-image translation}.
In: \bbtitle{Proceedings of the European Conference on Computer Vision (ECCV)},
pp. \bfpage{172}--\blpage{189}
(\byear{2018}).
\doiurl{10.1007/978-3-030-01219-9_11}
\end{bchapter}
\endbibitem

\bibitem{zhang2020}
\begin{barticle}
\bauthor{\bsnm{Zhang}, \binits{F.}},
\bauthor{\bsnm{Gao}, \binits{H.}},
\bauthor{\bsnm{Lai}, \binits{Y.}}:
\batitle{Detail-preserving cyclegan-adain framework for image-to-ink painting translation}.
\bjtitle{IEEE Access}
\bvolume{8},
\bfpage{132002}--\blpage{132011}
(\byear{2020}).
\doiurl{10.1109/access.2020.3009470}
\end{barticle}
\endbibitem

\bibitem{chung2022}
\begin{botherref}
\oauthor{\bsnm{Chung}, \binits{C.-Y.}},
\oauthor{\bsnm{Huang}, \binits{S.-H.}}:
Interactively transforming chinese ink paintings into realistic images using a border enhance generative adversarial network.
Multimedia Tools and Applications,
1--34
(2022).
\doiurl{10.1007/s11042-022-13684-4}
\end{botherref}
\endbibitem

\bibitem{he2016deep}
\begin{bchapter}
\bauthor{\bsnm{He}, \binits{K.}},
\bauthor{\bsnm{Zhang}, \binits{X.}},
\bauthor{\bsnm{Ren}, \binits{S.}},
\bauthor{\bsnm{Sun}, \binits{J.}}:
\bctitle{Deep residual learning for image recognition}.
In: \bbtitle{Proceedings of the IEEE Conference on Computer Vision and Pattern Recognition},
pp. \bfpage{770}--\blpage{778}
(\byear{2016}).
\doiurl{10.1109/cvpr.2016.90}
\end{bchapter}
\endbibitem

\bibitem{huang2017densely}
\begin{bchapter}
\bauthor{\bsnm{Huang}, \binits{G.}},
\bauthor{\bsnm{Liu}, \binits{Z.}},
\bauthor{\bsnm{Van Der~Maaten}, \binits{L.}},
\bauthor{\bsnm{Weinberger}, \binits{K.Q.}}:
\bctitle{Densely connected convolutional networks}.
In: \bbtitle{Proceedings of the IEEE Conference on Computer Vision and Pattern Recognition},
pp. \bfpage{4700}--\blpage{4708}
(\byear{2017}).
\doiurl{10.1109/cvpr.2017.243}
\end{bchapter}
\endbibitem

\bibitem{mao2017}
\begin{bchapter}
\bauthor{\bsnm{Mao}, \binits{X.}},
\bauthor{\bsnm{Li}, \binits{Q.}},
\bauthor{\bsnm{Xie}, \binits{H.}},
\bauthor{\bsnm{Lau}, \binits{R.Y.}},
\bauthor{\bsnm{Wang}, \binits{Z.}},
\bauthor{\bsnm{Paul~Smolley}, \binits{S.}}:
\bctitle{Least squares generative adversarial networks}.
In: \bbtitle{Proceedings of the IEEE International Conference on Computer Vision},
pp. \bfpage{2794}--\blpage{2802}
(\byear{2017}).
\doiurl{10.1109/iccv.2017.304}
\end{bchapter}
\endbibitem

\bibitem{simonyan2014}
\begin{barticle}
\bauthor{\bsnm{Simonyan}, \binits{K.}},
\bauthor{\bsnm{Zisserman}, \binits{A.}}:
\batitle{Very deep convolutional networks for large-scale image recognition}.
\bjtitle{arXiv preprint arXiv:1409.1556}
(\byear{2014}).
\doiurl{10.48550/arXiv.1409.1556}
\end{barticle}
\endbibitem

\bibitem{poma2020}
\begin{bchapter}
\bauthor{\bsnm{Poma}, \binits{X.S.}},
\bauthor{\bsnm{Riba}, \binits{E.}},
\bauthor{\bsnm{Sappa}, \binits{A.}}:
\bctitle{Dense extreme inception network: Towards a robust cnn model for edge detection}.
In: \bbtitle{Proceedings of the IEEE/CVF Winter Conference on Applications of Computer Vision},
pp. \bfpage{1923}--\blpage{1932}
(\byear{2020}).
\doiurl{10.1109/wacv45572.2020.9093290}
\end{bchapter}
\endbibitem

\bibitem{zhang2018}
\begin{bchapter}
\bauthor{\bsnm{Zhang}, \binits{R.}},
\bauthor{\bsnm{Isola}, \binits{P.}},
\bauthor{\bsnm{Efros}, \binits{A.A.}},
\bauthor{\bsnm{Shechtman}, \binits{E.}},
\bauthor{\bsnm{Wang}, \binits{O.}}:
\bctitle{The unreasonable effectiveness of deep features as a perceptual metric}.
In: \bbtitle{Proceedings of the IEEE Conference on Computer Vision and Pattern Recognition},
pp. \bfpage{586}--\blpage{595}
(\byear{2018}).
\doiurl{10.1109/cvpr.2018.00068}
\end{bchapter}
\endbibitem

\bibitem{paszke2019}
\begin{botherref}
\oauthor{\bsnm{Paszke}, \binits{A.}},
\oauthor{\bsnm{Gross}, \binits{S.}},
\oauthor{\bsnm{Massa}, \binits{F.}},
\oauthor{\bsnm{Lerer}, \binits{A.}},
\oauthor{\bsnm{Bradbury}, \binits{J.}},
\oauthor{\bsnm{Chanan}, \binits{G.}},
\oauthor{\bsnm{Killeen}, \binits{T.}},
\oauthor{\bsnm{Lin}, \binits{Z.}},
\oauthor{\bsnm{Gimelshein}, \binits{N.}},
\oauthor{\bsnm{Antiga}, \binits{L.}}:
Pytorch: An imperative style, high-performance deep learning library.
Advances in neural information processing systems
\textbf{32}
(2019)
\end{botherref}
\endbibitem

\bibitem{kingma2014}
\begin{barticle}
\bauthor{\bsnm{Kingma}, \binits{D.P.}},
\bauthor{\bsnm{Ba}, \binits{J.}}:
\batitle{Adam: A method for stochastic optimization}.
\bjtitle{arXiv preprint arXiv:1412.6980}
(\byear{2014}).
\doiurl{10.48550/arXiv.1412.6980}
\end{barticle}
\endbibitem

\bibitem{huang2017}
\begin{bchapter}
\bauthor{\bsnm{Huang}, \binits{X.}},
\bauthor{\bsnm{Belongie}, \binits{S.}}:
\bctitle{Arbitrary style transfer in real-time with adaptive instance normalization}.
In: \bbtitle{Proceedings of the IEEE International Conference on Computer Vision},
pp. \bfpage{1501}--\blpage{1510}
(\byear{2017}).
\doiurl{10.1109/iccv.2017.167}
\end{bchapter}
\endbibitem

\bibitem{dou2020}
\begin{barticle}
\bauthor{\bsnm{Dou}, \binits{H.}},
\bauthor{\bsnm{Chen}, \binits{C.}},
\bauthor{\bsnm{Hu}, \binits{X.}},
\bauthor{\bsnm{Jia}, \binits{L.}},
\bauthor{\bsnm{Peng}, \binits{S.}}:
\batitle{Asymmetric cyclegan for image-to-image translations with uneven complexities}.
\bjtitle{Neurocomputing}
\bvolume{415},
\bfpage{114}--\blpage{122}
(\byear{2020}).
\doiurl{10.1016/j.neucom.2020.07.044}
\end{barticle}
\endbibitem

\bibitem{peng2023unsupervised}
\begin{barticle}
\bauthor{\bsnm{Peng}, \binits{Z.}},
\bauthor{\bsnm{Wang}, \binits{H.}},
\bauthor{\bsnm{Weng}, \binits{Y.}},
\bauthor{\bsnm{Yang}, \binits{Y.}},
\bauthor{\bsnm{Shao}, \binits{T.}}:
\batitle{Unsupervised image translation with distributional semantics awareness}.
\bjtitle{Computational Visual Media}
\bvolume{9}(\bissue{3}),
\bfpage{619}--\blpage{631}
(\byear{2023}).
\doiurl{10.1007/s41095-022-0295-3}
\end{barticle}
\endbibitem

\bibitem{liu2017unsupervised}
\begin{botherref}
\oauthor{\bsnm{Liu}, \binits{M.-Y.}},
\oauthor{\bsnm{Breuel}, \binits{T.}},
\oauthor{\bsnm{Kautz}, \binits{J.}}:
Unsupervised image-to-image translation networks.
Advances in neural information processing systems
\textbf{30}
(2017)
\end{botherref}
\endbibitem

\bibitem{tang2021attentiongan}
\begin{barticle}
\bauthor{\bsnm{Tang}, \binits{H.}},
\bauthor{\bsnm{Liu}, \binits{H.}},
\bauthor{\bsnm{Xu}, \binits{D.}},
\bauthor{\bsnm{Torr}, \binits{P.H.}},
\bauthor{\bsnm{Sebe}, \binits{N.}}:
\batitle{Attentiongan: Unpaired image-to-image translation using attention-guided generative adversarial networks}.
\bjtitle{IEEE transactions on neural networks and learning systems}
(\byear{2021}).
\doiurl{10.1109/TNNLS.2021.3105725}
\end{barticle}
\endbibitem

\bibitem{heusel2017}
\begin{botherref}
\oauthor{\bsnm{Heusel}, \binits{M.}},
\oauthor{\bsnm{Ramsauer}, \binits{H.}},
\oauthor{\bsnm{Unterthiner}, \binits{T.}},
\oauthor{\bsnm{Nessler}, \binits{B.}},
\oauthor{\bsnm{Hochreiter}, \binits{S.}}:
Gans trained by a two time-scale update rule converge to a local nash equilibrium.
Advances in neural information processing systems
\textbf{30}
(2017)
\end{botherref}
\endbibitem

\bibitem{binkowski2018}
\begin{barticle}
\bauthor{\bsnm{Bi{\'n}kowski}, \binits{M.}},
\bauthor{\bsnm{Sutherland}, \binits{D.J.}},
\bauthor{\bsnm{Arbel}, \binits{M.}},
\bauthor{\bsnm{Gretton}, \binits{A.}}:
\batitle{Demystifying mmd gans}.
\bjtitle{arXiv preprint arXiv:1801.01401}
(\byear{2018}).
\doiurl{10.48550/arXiv.1801.01401}
\end{barticle}
\endbibitem

\bibitem{hore2010}
\begin{bchapter}
\bauthor{\bsnm{Hore}, \binits{A.}},
\bauthor{\bsnm{Ziou}, \binits{D.}}:
\bctitle{Image quality metrics: Psnr vs. ssim}.
In: \bbtitle{2010 20th International Conference on Pattern Recognition},
pp. \bfpage{2366}--\blpage{2369}
(\byear{2010}).
\doiurl{10.1109/icpr.2010.579}.
\bcomment{IEEE}
\end{bchapter}
\endbibitem

\end{thebibliography}


\end{document}